%% file: neurips_2025.tex
\documentclass{article}

\usepackage[main, final]{neurips_2025}
\usepackage{graphicx}
\usepackage{subfigure}
\usepackage{algorithm}
\usepackage{algorithmicx}
\usepackage{algpseudocode}
\usepackage{amsmath}
\usepackage{hyperref}
\usepackage{pifont}
\usepackage{caption}

\usepackage[table,xcdraw]{xcolor}
\usepackage{caption}

\usepackage{makecell}
\usepackage[utf8]{inputenc} % allow utf-8 input
\usepackage[T1]{fontenc}    % use 8-bit T1 fonts
\usepackage{booktabs}       % professional-quality tables
\usepackage{amsfonts}       % blackboard math symbols
\usepackage{nicefrac}       % compact symbols for 1/2, etc.
\usepackage{microtype}      % microtypography
\usepackage{tabularx}

\newcommand{\cmark}{\textcolor{green!60!black}{\ding{51}}} % 对号 √
\newcommand{\xmark}{\textcolor{red!70!black}{\ding{55}}}   % 叉号 ×

\definecolor{best}{HTML}{F1A3A3}
\definecolor{second}{HTML}{F6CF88}
\definecolor{third}{HTML}{FFFC9E}

\usepackage[utf8]{inputenc} % allow utf-8 input
\usepackage[T1]{fontenc}    % use 8-bit T1 fonts
\usepackage{hyperref}       % hyperlinks
\usepackage{url}            % simple URL typesetting
\usepackage{booktabs}       % professional-quality tables
\usepackage{amsfonts}       % blackboard math symbols
\usepackage{nicefrac}       % compact symbols for 1/2, etc.
\usepackage{microtype}      % microtypography
\usepackage{xcolor}         % colors

\newcommand{\numcite}[1]{[\citenum{#1}]}

\title{H3D-DGS: Exploring Heterogeneous 3D Motion Representation for Deformable 3D Gaussian Splatting}

\author{Bing He 
\textsuperscript{1}
\thanks{Authors contributed equally to this work.} 
\& Yunuo Chen
\textsuperscript{1}
\footnotemark[1]
\& Guo Lu 
\textsuperscript{1}
\& Qi Wang
\textsuperscript{2}
\& Qunshan Gu 
\textsuperscript{2} \\
\textbf{\& Rong Xie}
\textsuperscript{1}
\textbf{\& Li Song
\textsuperscript{1}
\thanks{Corresponding author.}
\& Wenjun Zhang 
\textsuperscript{1}}
\\
\textsuperscript{1} Shanghai Jiao Tong University,   \textsuperscript{2} Alibaba Group \\
\texttt{\{sandwich\_theorem\}@sjtu.edu.cn} \\
\texttt{\{cyril-chenyn\}@sjtu.edu.cn} \\
}

\begin{document}

\maketitle

\includegraphics[width=\textwidth]{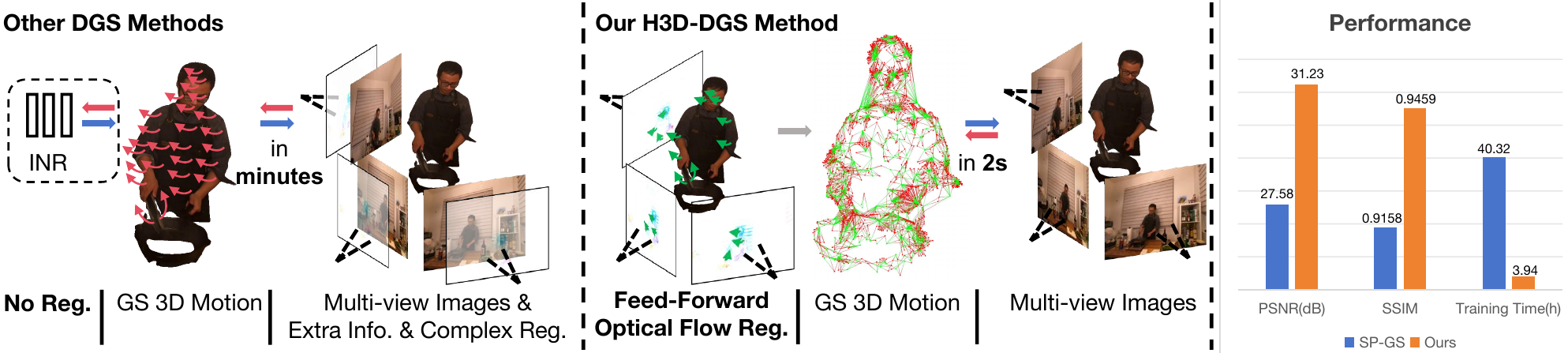}
% \captionof{figure}{Caption of the teaser }
% \vspace{-2em}
\captionof{figure}{\textbf{H3D-DGS}: Previous deformable 3DGS methods heavily depend on gradient-based methods, often introducing complex loss regularizations to recover 3D motion. 
We challenge this convention by observing that, the observability of 3D motion varies across spatial directions. 
Specifically, the motion components observable in camera’s image plane can be directly estimated using optical flow.
In contrast, motion components orthogonal to the image plane must be inferred from additional viewpoints. 
We propose to directly incorporate these observable components into a fixed, feed-forward 3D motion representation, allowing the neural network to focus solely on learning the unobservable motion.
To this end, we introduce a heterogeneous 3D motion representation—\textbf{H3D control points}—which decouple observable and learnable motion components.
This heterogeneous structure serves as an effective inductive bias, promoting physically plausible 3D motion estimation.
our H3D-DGS achieves both superior performance and faster convergence compared to existing methods.
}
\label{fig:teaser}

\input{sec/0_abstract}    
\input{sec/1_intro}
\input{sec/2_relatedwork}

\input{sec/3_method}

\input{sec/4_experiment}
\input{sec/5_conclusion}

{
    \small
    \bibliographystyle{ieeenat_fullname}
    \bibliography{neurips_2025}
}

\newpage
\input{sec/X_suppl}

\end{document}

%% file: sec/0_abstract.tex
\begin{abstract}
Dynamic scene reconstruction poses a persistent challenge in 3D vision. Deformable 3D Gaussian Splatting has emerged as an effective method for this task, offering real-time rendering and high visual fidelity.
This approach decomposes a dynamic scene into a static representation in a canonical space and time-varying scene motion.
Scene motion is defined as the collective movement of all Gaussian points, and for compactness, existing approaches commonly adopt implicit neural fields or sparse control points. 
However, these methods predominantly rely on gradient-based optimization for all motion information. Due to the high degree of freedom, they struggle to converge on real-world datasets exhibiting complex motion.
To preserve the compactness of motion representation and address convergence challenges, this paper proposes heterogeneous 3D control points, termed \textbf{H3D control points}, whose attributes are obtained using a hybrid strategy combining optical flow back-projection and gradient-based methods. 
This design decouples directly observable motion components from those that are geometrically occluded.
Specifically, components of 3D motion that project onto the image plane are directly acquired via optical flow back projection, while unobservable portions are refined through gradient-based optimization.
Experiments on the Neu3DV and CMU-Panoptic datasets demonstrate that our method achieves superior performance over state-of-the-art deformable 3D Gaussian splatting techniques. Remarkably, our method converges within just 100 iterations and achieves a per-frame processing speed of 2 seconds on a single NVIDIA RTX 4070 GPU.
\end{abstract}

% Mainstream approaches typically employ a global deformation field to warp a 3D scene in canonical space. However, the inherent low-frequency nature of implicit neural fields often leads to ineffective representations of complex motions. Moreover, their structural rigidity can hinder adaptation to scenes with varying resolutions and durations.
% To address these challenges, we introduce a novel approach for streaming 4D real-world reconstruction utilizing discrete 3D control points. This method physically models local rays and establishes a motion-decoupling coordinate system.

%% file: sec/1_intro.tex
\section{Introduction}

Reconstructing real-world scenes is a long-standing challenge in the field of 3D vision. Recently, 3D Gaussian Splatting (3D-GS)~\cite{kerbl20233d} has demonstrated remarkable success in producing high-quality reconstructions for static scenes. This technique utilizes Gaussians to model the scene, assigning them with physically meaningful properties, and renders image by "splatting" these Gaussians onto the image plane. 
\label{discrete}

\textbf{Why control points?} Compared to NeRF~\cite{mildenhall2021nerf}, a key advantage of 3D-GS lies in its discrete structure. This structure ensures that scene representation—via Gaussians—is concentrated at occupied regions within the scene, avoiding the inefficiencies of a global field that allocates unnecessary representational capacity to empty space.
Similarly, the global implicit neural field used for deformation faces a similar issue: only a small part of the scene is dynamic, while the majority remains static. Therefore, a discrete, localized motion representation holds promise, as it offers precise and flexible modeling of 3D motions at a local level. 

In our approach, we adopt control points as the motion representation because, despite being compact, they can efficiently represent local motion. We further distinguish between the background and the moving objects in the 3D scene, such that our control points are applied only to the latter.

\textbf{Why heterogeneous?} Realizing accurate 3D motion representation requires reconstructing motion unobservable from a single camera. In a multi-view system, traditional graphics methods~\cite{vedula1999three} struggle to align complex 3D points. While gradient-based methods~\cite{luiten2023dynamic} circumvent the 3D alignment problem, they suffer from poor convergence due to their high degrees of freedom (DoF) and require extensive regularization, making them time-consuming. 

Our heterogeneous approach differs by combining graphics-based techniques with gradient-based methods. 
Recognizing that optical flow is the 2D projection of scene flow, we introduce a local decoupling strategy, dividing local 3D motion into observable and unobservable components. Components of 3D motion that project onto the camera plane are directly acquired via optical flow back projection. In contrast, the unobservable portion is refined through gradient-based methods. Since we have incorporated regularization into the structural design of the control point, backpropagation in our method focuses only on the complex portion of the 3D motion. Our method mitigates convergence issues and achieves fast optimization.
This new form of heterogeneous 3D motion representation is referred to as the "H3D control points" approach.

\textbf{Streaming framework.} Building on the introduced 3D motion representation, we present a novel generalized streaming framework for dynamic 3D real-world reconstruction with multiple camera setups. Beginning with an initial 3D reconstruction, our workflow decomposes the dynamic reconstruction process into distinct submodules: 3D segmentation, H3D control point generation, object-wise motion manipulation, and residual compensation. This structured approach minimizes accumulated errors and ensures a compact and robust representation. 

Our key contributions are as follows:
\begin{itemize}
\item We propose a strategy to split 3D motion into observable and unobservable components. Observable motion (projected to the camera plane) is estimated directly via optical flow back-projection, while unobservable components are optimized using gradients backpropagation.
\item We introduce \textbf{H3D control points}, a heterogeneous motion representation that significantly improves convergence and accuracy. Our method converges within 100 iterations and achieves a processing speed of 2 seconds per frame on a single NVIDIA 4070 GPU.
\item We develop a streaming framework for real-world dynamic scene reconstruction, setting a new benchmark on the Neu3DV and CMU-Panoptic datasets.

\end{itemize}

%% file: sec/2_relatedwork.tex
\section{Related Work}
\subsection{Dynamic Scene Reconstruction}

Neural Radiance Field (NeRF)~\cite{mildenhall2021nerf} have demonstrated strong performance in novel view synthesis by modeling scenes with global continuous implicit functions.
Numerous extensions have adapted NeRF to dynamic scenes. Mainstream methods~\cite{xian2021space,wang2021neural,park2021nerfies,park2021hypernerf,pumarola2021d,du2021neural,li2021neural,liu2022devrf,fang2022fast,shao2023tensor4d,cao2023hexplane,fridovich2023k,liu2023robust,song2023nerfplayer} modeled dynamic scenes by learning a deformation field that warps a canonical 3D representation over time.
Other research~\cite{wang2023mixed,li2023dynibar,lin2022efficient,lin2023im4d} improved reconstruction quality by incorporating camera pose priors. Additional supervision information, such as depth~\cite{attal2021torf} and optical flow~\cite{wang2023flow}, were also employed to guide training. NeRFPlayer~\cite{song2023nerfplayer}, used self-supervised learning to segment dynamic scenes into static, deforming, and newly appearing regions, applying tailored strategies to each.

Recently, 3D-GS~\cite{kerbl20233d} introduced an elegant point-based rendering approach with efficient CUDA implementations. Many 4D Gaussian methods parallel NeRF-based approaches by incorporating temporal dynamics into spatial representations. For instance, \citet{yang2023real} incorporated time-variant attributes into Gaussians, while Dynamic-GS~\cite{luiten2023dynamic} learned dense Gaussian motion directly. Subsequent works~\cite{guo2024motion,zhu2024motiongs} incorporated optical flow to enhance motion accuracy. 3DGStream~\cite{sun20243dgstream} proposed a Neural Transformation Cache to model per-frame motion. Gaussian-Flow~\cite{lin2024gaussian} introduced a Dual-Domain Deformation Model for point-wise motion representation. Spacetime Gaussian~\cite{li2024spacetime} proposed a feature splatting and rendering approach. Several studies~\cite{wu20234d,yang2023deformable,huang2023sc,lin2024gaussian,das2024neural,yu2024cogs} deformed canonical Gaussians using global implicit fields to capture 4D dynamics.

\subsection{3D Control Points}

Real-world scenes typically consist of a large static background and a smaller dynamic foreground. 
While global neural fields are compact in representation, they often lack the flexibility to selectively model only the dynamic regions, and require architectural redesigns to adapt across different scenes. 
In contrast, 3D control point methods are promising due to their ability to flexibly capture localized motion and scale effectively.
Traditional graphics approaches~\cite{sorkine2005laplacian,yu2004mesh} have long offered flexible deformation techniques that preserve geometric details. Among these, \citet{sumner2007embedded} introduced Embedded Deformation (ED) graphs, which use sparse control points to represent the motion of dense surfaces, achieving a balance between compactness and flexibility. HiFi4G~\cite{jiang2023hifi4g} directly adopted ED-graphs, but its surface reconstruction requires dense camera coverage and is computationally expensive. Moreover, unlike meshes, the volumetric radiance representation of 3D Gaussians is not well-suited for the thin nature of surfaces~\cite{huang20242d}.
Recent gradient-based methods have explored the use of control points for compact motion representation in 3D. 
SP-GS~\cite{wan2024superpoint} clustered Gaussians into superpoints, primitive motion groups, while SC-GS~\cite{huang2023sc} also adopted the concept of control points, although its direct optimization is challenged by high degrees of freedom (DoFs).

\begin{figure}[H]
\centering
\begin{minipage}[t]{0.4\textwidth}
\small
\fontsize{7}{8}\selectfont
\setlength{\tabcolsep}{8pt} % 设置列间距为 8pt
\centering
\captionof{table}{Method Comparison. Rep. and Manip. are abbreviations for representation and manipulation, respectively.}

\begin{tabular}{ccccc}
\toprule
      & \begin{tabular}[c]{@{}c@{}}Large\\ Move\end{tabular} 
      & \begin{tabular}[c]{@{}c@{}}Fast\\ Train\end{tabular} 
      & \begin{tabular}[c]{@{}c@{}}Compact\\ Rep.\end{tabular} 
      & \begin{tabular}[c]{@{}c@{}}Motion\\ Manip.\end{tabular} \\
\midrule
Dy-GS & \cmark & \xmark & \xmark & \xmark \\
MA-GS & \xmark & \xmark & \cmark & \xmark \\
4D-GS & \xmark & \cmark & \cmark & \xmark \\
SP-GS & \xmark & \xmark & \cmark & \xmark \\
SC-GS & \xmark & \xmark & \cmark & \cmark \\
\midrule
Ours  & \cmark & \cmark & \cmark & \cmark \\
\bottomrule
\end{tabular}
\label{tab:example}
\end{minipage}
\hfill
\begin{minipage}[t]{0.5\textwidth}
\centering
% \linewidth
\captionof{table}{Average reconstruction results for the Neu3DV dataset. Training time is reported in hours. }
\fontsize{7}{9.2} \selectfont
\setlength{\tabcolsep}{9pt} % 设置列间距为 5pt
\begin{tabularx}{\linewidth}{lccccccccccc}
\toprule
  \begin{tabular}[c]{@{}c@{}}Metrics\end{tabular} 
& \begin{tabular}[c]{@{}c@{}}PSNR$\uparrow$\end{tabular} 
& \begin{tabular}[c]{@{}c@{}}SSIM$\uparrow$\end{tabular} 
& \begin{tabular}[c]{@{}c@{}}LPIPS$\downarrow$\end{tabular} 
& \begin{tabular}[c]{@{}c@{}}Training\\ Time.\end{tabular} \\

\midrule
% \multicolumn{3}{l}{Dy-GS}
Dy-GS & 27.65                         & 0.9232                         & 0.1313                         & 57.35                        \\
% \multicolumn{3}{l}{MA-GS}
MA-GS & 28.76                         & 0.9299                         & 0.1146                         & 45.38                        \\
% \multicolumn{3}{l}{4D-GS}
4D-GS & 30.49                         & 0.9401                        & 0.0998                         & 6.88                                    \\
% \multicolumn{3}{l}{SP-GS}
SP-GS & 27.03                         & 0.9109                        & 0.1225                         & 40.32                                    \\

\midrule
% \multicolumn{3}{l}{Ours} 
Ours & \textbf{30.91} & \textbf{0.9437} & \textbf{0.0941} & \textbf{1.89}       \\
\bottomrule
\end{tabularx}
\label{Average}
\end{minipage}
\end{figure}

\begin{figure*}[ht]
  \centering
  \includegraphics[width=1\linewidth]{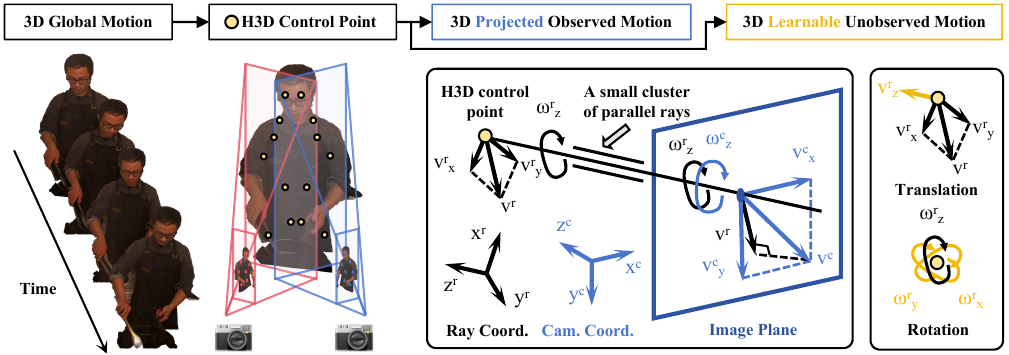}
  \caption{\textbf{H3D Control Points.} To predict dense 3D motion in a sparse manner, we propose H3D control points containing local translation and rotation information. Unlike previous works which learn all motion information with gradient-based method, we exploit 2D motion priors derived from the optical flow. Both translation and rotation are divided into projected observable part and learnable unobservable part. We model the localized light as parallel rays and make a detailed derivation.
}
 \label{H3D}
 \vspace{-10pt}
\end{figure*}
Our method introduces four key distinctions from existing gradient-based approaches. First, our method robustly handles real-world datasets with large-scale motion~\cite{Joo_2017_TPAMI}, where prior methods~\cite{guo2024motion,huang2023sc,wan2024superpoint,wu20234d} struggle. Second, we introduce structural regularization that improves training efficiency, enabling our model to train as fast as 2 seconds per frame. Third, our H3D control points offer a compact motion representation, requiring only 0.2\% as many control points as Gaussians. In comparison, methods like Dynamic-GS\cite{luiten2023dynamic} and GS-Flow~\cite{gao2024gaussianflow} use redundant motion representations, applying residuals to every Gaussian.  Fourth, our control points influence nearby Gaussians spatially, inheriting the desirable properties of traditional control points and supporting user-driven motion manipulation~\cite{huang2023sc}.

%% file: sec/3_method.tex
\section{Method}

In this section, we introduce our method for dynamic scene reconstruction with Gaussians. First, we present the notations and symbols used throughout the section to facilitate understanding. We then begin with the key idea of our approach—motion decoupling—and provide its detailed formulation. Next, we introduce the control point action pattern, where motion information is constructed using the proposed motion decoupling strategy. Based on the resulting H3D control points, we describe the full pipeline for dynamic scene reconstruction. Finally, we present the associated loss functions used to optimize the model.

\subsection{Mathematical Notation}
The symbols used in our method are defined as follows: 

Scalar values are denoted in standard font, while vectors and matrices are bolded. The superscripts indicate reference systems, and the subscripts denote point-specific information. Euclidean distance is represented by $\|\cdot\|$, and learnable parameters are marked with a hat symbol $\hat{\cdot}$.

Positions $x, y$ and optical flow values $u, v$ are measured in pixels, while 3D positions $X, Y, Z$ and velocities $\mathbf{V}$ are measured in meters. The focal length of the camera is denoted as $f$, in pixels in Eq.~\ref{pos}, in meters in Eqs.~\ref{flow}~\ref{trans}, $(c_x, c_y)$ indicates the center of projection in pixels, $\mathbf{K}$ represents the intrinsic matrix of the camera, $\mathbf{R}$ is the rotation matrix, $\mathbf{q}$ denotes the unit rotation quaternion, and $\mathbf{t}$ represents translation.

For simplicity, linear velocity $\mathbf{V}$ and translation $\mathbf{t}$, as well as angular velocity $\boldsymbol{\omega}$ and angular variation represented by the unit quaternion $\mathbf{q}$, are treated analogously in our calculations, as both represent changes in motion attributes between two neighboring frames.

\begin{figure*}[ht]
  \centering
   \subfigure[]{
    \includegraphics[width=0.26\textwidth]{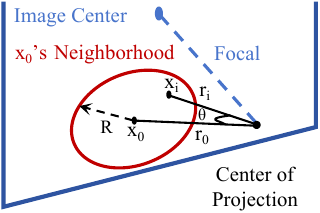}
    \label{fig:parallel}
    }
    % \hspace{0.03\textwidth}
    \subfigure[]{
    \includegraphics[width=0.26\textwidth]{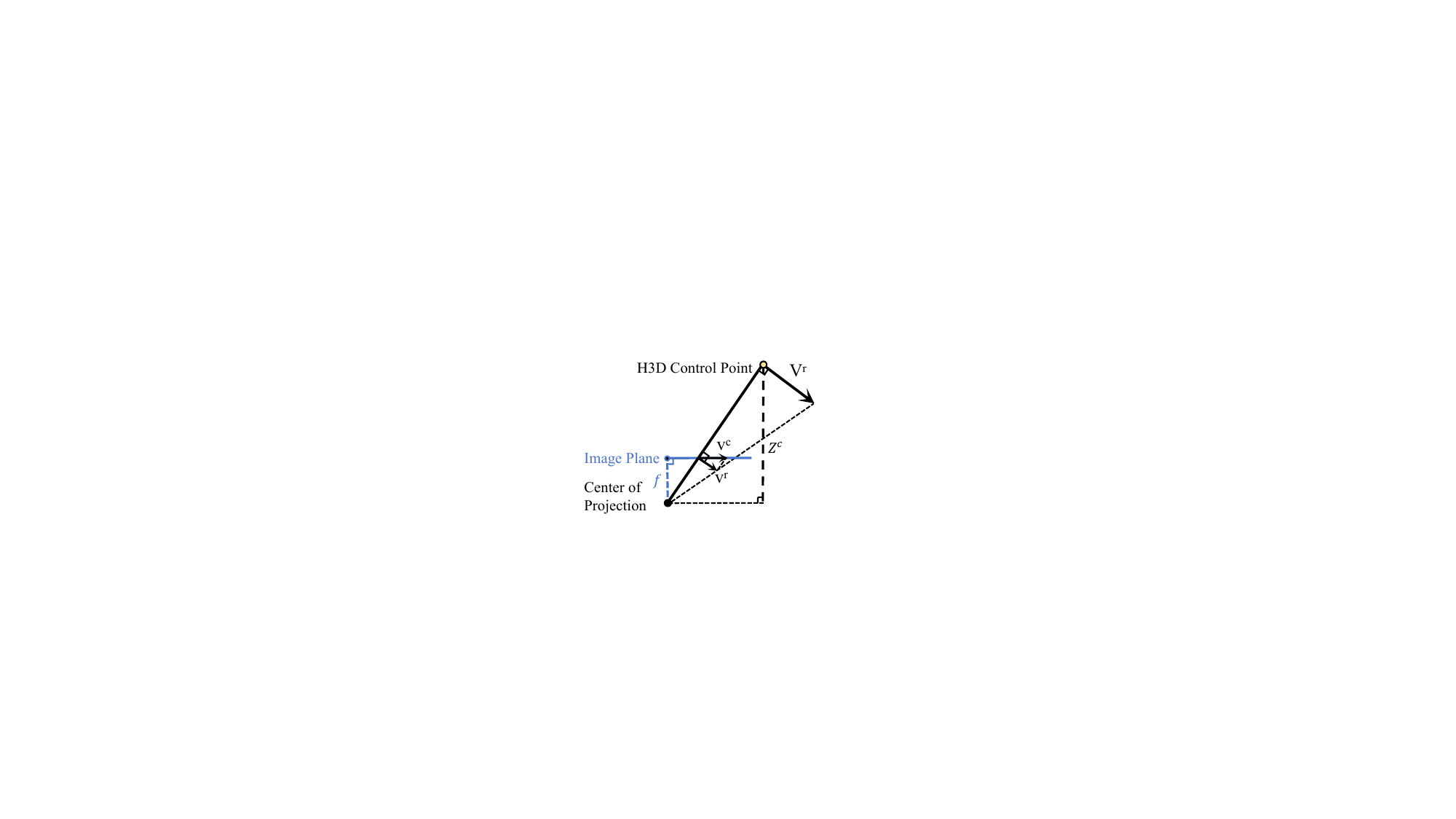}
    \label{fig:proj}
  }
  % \hspace{0.03\textwidth}
    \subfigure[]{
    \includegraphics[width=0.28\textwidth]{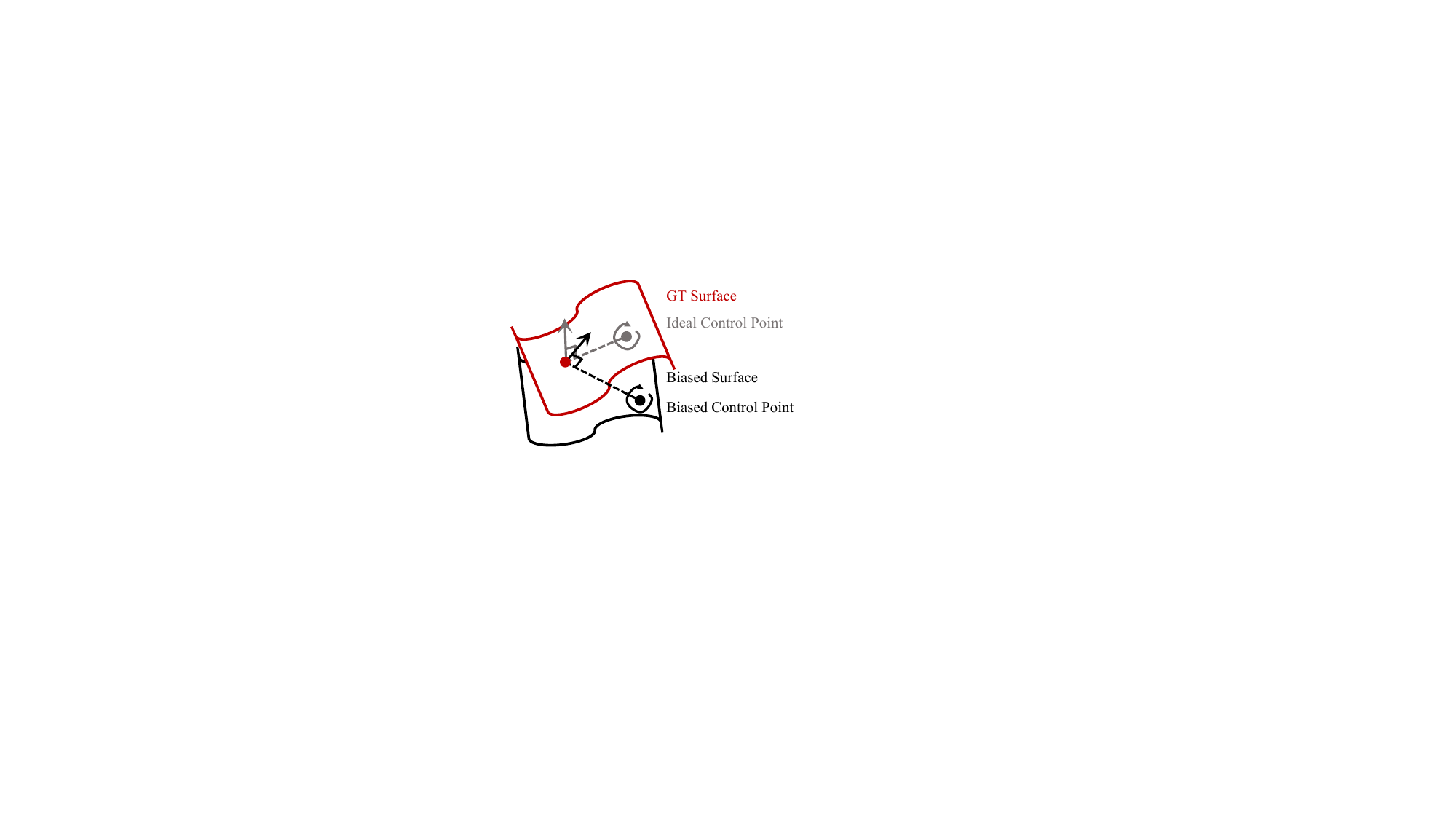}
    \label{fig:bias}
  }
  \subfigure[]{
    \includegraphics[width=0.13\textwidth]{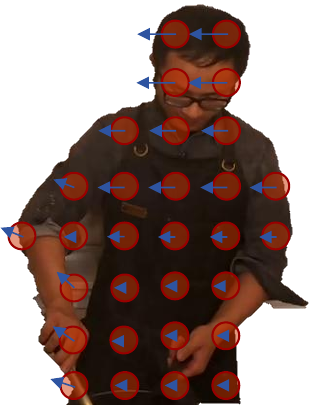}
    \label{fig:grid}
  }
  \caption{\textbf{Auxiliary Diagrams for Local Motion Mapping.} (a) An illustration of angles, points, and rays within the neighborhood of \(\mathbf{x}_0\). (b) A quantitative depiction of motion projection. (c) A comparison between Gaussians distributed on the ground truth surface and control points located on a biased surface. (d) Illustration for grid sampling. Grid sampling is performed independently for each camera to provide a 2D motion prior for H3D control points.}
  \vspace{-10pt}
\end{figure*}

\subsection{Local 3D Motion Decoupling and Heterogeneous 3D Control Points Generation }
\label{Control}

To effectively represent local translation and rotation, the attributes of a 3D control point encompass its 3D position, local spatial translation, and rotation.

Typically, the 3D position of a point is obtained by back projection. Given a point located at $(x^c_0,y^c_0)$ on the image plane with depth $Z^c$, its 3D position in the camera coordinate system is defined as:

\begin{equation}
\mathbf{X}^c=\frac{Z^c}{f}(x^c_0-c_x,y^c_0-c_y,f)^{\mathrm{T}}.
\label{pos}
\end{equation}

Now we focus on translation and rotation. At any given viewpoint, a portion of the 3D rigid motion is projected onto the image plane. This motion component can therefore be extracted directly from optical flow and projected into space with appropriate design.

To facilitate this, we define a "ray coordinate system," where the z-axis aligns with the H3D control point, extending from the camera toward the scene. This motion decoupling process is illustrated in Fig.~\ref{H3D}. The components $v^r_x, v^r_y, {\omega}^r_{z}$ are derived from optical flow, while $v^r_z, {\omega}^r_{x}, {\omega}^r_{y}$ remain learnable.

To extend the motion representation to the local space rather than just a single point, we introduce a local ray approximation assumption. This allows any point $x_i$ in the neighborhood \(\mathcal{N}\) around $x_0$ (see Fig.~\ref{fig:parallel}) to share the same local coordinate system. The derivation, given in the Appendix, demonstrates that parallel light retains only motion components perpendicular to the ray direction. The relationship between image plane optical flow and true motion information is thus projective:

\begin{equation}
\mathbf{v}^r=f*\frac{1}{N}\sum_{\mathbf{x}_i \in \mathcal{N}}(\frac{\mathrm{d} x^r_{i}}{\mathrm{d} t},\frac{\mathrm{d} y^r_{i}}{\mathrm{d} t})^{\mathrm{T}}
=f*\frac{1}{N}\left[\mathbf{R}_{cr}\right]_{:2,:2}\left[\mathbf{K^{-1}}\right]_{:2,:2}\sum_{\mathbf{x}_i \in \mathcal{N}}(u_i,v_i)^{\mathrm{T}}
\label{flow}
\end{equation}

where \((u_i, v_i)\) denotes the optical flow of points \(\mathbf{x}_i\). The inverse intrinsic matrix \(\mathbf{K^{-1}}\) converts the optical flow to meters, and rotation matrix \(\mathbf{R}_{cr}\) transforms the system from the camera to the ray coordinate system. The notation \(\left[\cdot\right]_{:2,:2}\) selects the matrix's first two rows and columns, and \(N\) is the number of pixels in \(\mathcal{N}\). To better interpret scene dynamics \(\mathbf{v}_r\), we use the focal length \(f\) to adjust translation from normalized to physical space.

As illustrated in Fig.~\ref{fig:proj}, the formulas for translational and rotational information can be expressed respectively as:

\begin{equation}
    \mathbf{V}^r=(\frac{\mathrm{d} X^r}{\mathrm{d} t},\frac{\mathrm{d} Y^r}{\mathrm{d} t})^{\mathrm{T}}=\frac{Z^c}{f}\mathbf{v}^r,
\label{trans}
\end{equation}

\begin{equation}
{\omega}^r_{z}=\frac{1}{N}\sum_{\mathbf{x}_i \in \mathcal{N}}\frac{{(\mathbf{x}^r_{i}-\mathbf{x}^r_{0})} \times \mathbf{v}^r_{i}}{\left\|\mathbf{x}^r_{i}-\mathbf{x}^r_{0}\right\|^2}.
\label{angular}
\end{equation}

Thus, in the ray coordinate system, 3D position $\mathbf{X}^c$, 2D translation $\mathbf{V}^r$, and 1D rotation ${\omega}^r_{z}$ of the H3D control point are determined, while the learnable parts $\hat{V}^r_z, \hat{\omega}^r_x, \hat{\omega}^r_y$ complete the motion information, represented as:

\begin{equation}
    \mathbf{t^r}=(V^r_x,V^r_y,\hat{V}^r_z),
\end{equation}
\begin{equation}
    \mathbf{q^r}=\operatorname{Euler2Quat}((\hat{\omega}^r_x,\hat{\omega}^r_y,\omega^r_z)),
\end{equation}

The function $\operatorname{Euler2Quat}$ converts Euler angles to quaternions. Euler angles are more intuitive and interpretable for decoupling and controlling rotation components, while quaternions are used in the underlying 3D Gaussian Splatting framework operates to represent rotation.

At this stage, all attributes of the H3D control point are derived, completing its construction from the optical flow.
For each camera, H3D control points are independently sampled using a uniform grid pattern, as illustrated in Fig.\ref{fig:grid}. The sampling interval determines the density of H3D control points, and the radius of the circular sampling area represents the range of motion cues.
The H3D control points from all cameras are aggregated to represent the motion across the entire scene.
An intuitive schematic is also provided in Appendix Fig.~\ref{fig:ops}.

We also experimented with the incorporation of local macro-rotation following the ED-graph method~\cite{sumner2007embedded}, but observed a decline in performance. We attribute this to structural differences between Gaussians and meshes: while meshes are densely distributed along object surfaces, Gaussians exhibit a looser spatial distribution due to their higher degrees of freedom. Consequently, we omit the component concerning local macro rotation to minimize biases introduced by rough depth estimations, as illustrated in Fig.~\ref{fig:bias}. Although this adjustment compromises some sparsity, it optimizes our motion representation approach for volumetric radiance representations like Gaussians.

\subsection{Motion Manipulation}

The translation $\mathbf{t}$ and rotation $\mathbf{q}$ of each Gaussian $G$ are computed by interpolating from its K-nearest~\cite{cover1967nearest} control points $C_i$ within its spatial neighborhood $\mathcal{N}_G$. The interpolation weights $w_i$ are inversely proportional to the Euclidean distances to the control points.

\begin{equation}
    (\mathbf{t},\mathbf{q})=\sum_{C_i \in \mathcal{N}_G}w_{i}*(\mathbf{t}_{i}, \mathbf{q}_{i}),
\label{weighted}
\end{equation}

\begin{equation}
w_{i}=\frac{1/\left\|\mathbf{X}_{G}-\mathbf{X}_{{C}_i}\right\|}{\sum_{{{C}_i} \in \mathcal{N}_G} 1/\left\|\mathbf{X}_{G}-\mathbf{X}_{{C}_i}\right\|}.
\label{weighted2}
\end{equation}

The position $\mathbf{X}_t$ and rotation $\mathbf{q}_t$ of the Gaussian at time $t$ are given by:

\begin{equation}
    \mathbf{X}_t = \mathbf{t}+\mathbf{X}_{t-1}
\end{equation}
\begin{equation}
    \mathbf{q}_t = \mathbf{q}\mathbf{q}_{t-1}
\end{equation}

In practice, we set K = 3, meaning that each moving Gaussian derives its motion from the three nearest control points.

\subsection{Streaming workflow}

\begin{figure*}[ht]
  \centering
  \vspace{-10pt}
  \includegraphics[width=1\linewidth]{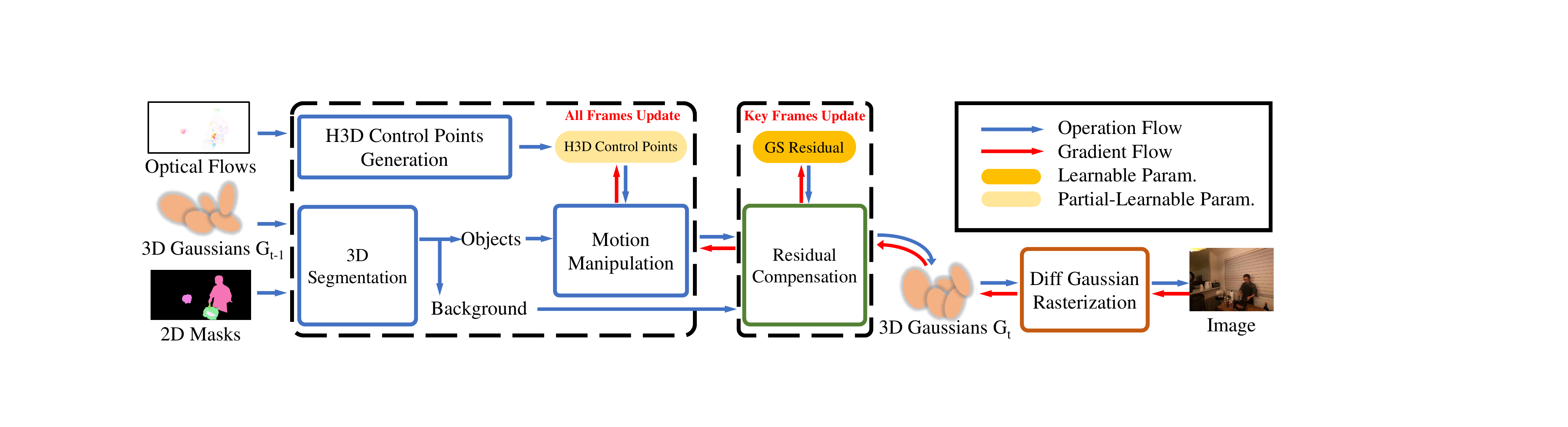}
  \caption{\textbf{Streaming workflow.} The workflow starts by segmenting the scene into a static background and moving objects using 3D segmentation algorithm. Optical flow is then applied to generate H3D control points. Motion-related attributes of the Gaussians are manipulated on an object-wise basis. To prevent reconstruction failures from error accumulation, Gaussian attributes are periodically updated in a keyframe manner, capturing additional scene information as attribute residuals of the Gaussians.}
  \label{fig:workflow}
  \vspace{-10pt}
\end{figure*}

The streaming workflow is depicted in Fig.~\ref{fig:workflow}. The proposed method comprises four independent modules: 3D segmentation, H3D control point generation, object-wise motion manipulation, and residual compensation. Inputs include 3D Gaussians, 2D masks, and optical flow. 3D Gaussians are obtained either from the initial static scene reconstruction or the previous frame. The 2D object masks are generated via the SAM-track method~\cite{cheng2023segment}. Optical flows are derived from the DIS method~\cite{kroeger2016fast}. 

\label{cate}

\textbf{3D Segmentation.} The purpose of 3D segmentation is to label each Gaussian as either part of a moving object or the static background. The control points act on their spatially proximate Gaussians. To ensure they only influence moving objects, it is essential to accurately define the spatial regions where each local representation applies.
To achieve this, we utilize multiview masks and employ a Gaussian category voting algorithm to segment the scene into dynamic objects and static background regions, following an approach similar to SA-GS~\cite{hu2024semantic}.
Specifically, each Gaussian is projected onto the image planes of all training viewpoints. By tallying the category labels across views, the Gaussian is assigned the category with the most votes. H3D control points, which also retain category labels when projected back into 3D, only manipulate Gaussians of the same category. This framework inherently supports topological transformations between objects with different categories but similar spatial locations.

\textbf{Residual Compensation.} This module is designed to mitigate error accumulation and maintain stable long-term reconstruction. Inspired by video coding techniques like residual coding~\cite{wiegand2003overview,sze2014high}, we integrate a keyframe-based update mechanism. 
The system performs full optimization of both Gaussians and H3D control points at keyframes, whereas during non-keyframe intervals, Gaussian attributes remain fixed and only control point parameters are updated. To formalize this structure, we introduce the concept of a Group of Scenes (GOS), where GOS-N denotes a sequence with one keyframe followed by N-1 non-keyframes. This design achieves a balance between computational efficiency and temporal reconstruction stability.

\subsection{Loss Function}

We adopt the same single-image reconstruction loss used in the original 3D Gaussian Splatting framework, consisting of an L1 loss combined with a D-SSIM term, which are computed between the rendered image $I_{render}$ and the ground truth image $I_{gt}$:
\begin{equation}
\begin{gathered}
    \mathcal{L} = (1-\lambda)\mathcal{L}_\mathrm{1}(I_{render},I_{gt}) + \lambda\mathcal{L}_\mathrm{D-SSIM}(I_{render},I_{gt}), \\
\end{gathered}   
\label{loss}
\end{equation}
with $\lambda$ set to 0.2. Our motion modeling and streaming pipeline are designed for both computational efficiency and robustness, enabling effective convergence without the need for additional loss constraints.

%% file: sec/4_experiment.tex
\section{Experiment}

\begin{figure*}[ht]
  \vspace{-10pt}
  \centering
  \includegraphics[width=1\linewidth]{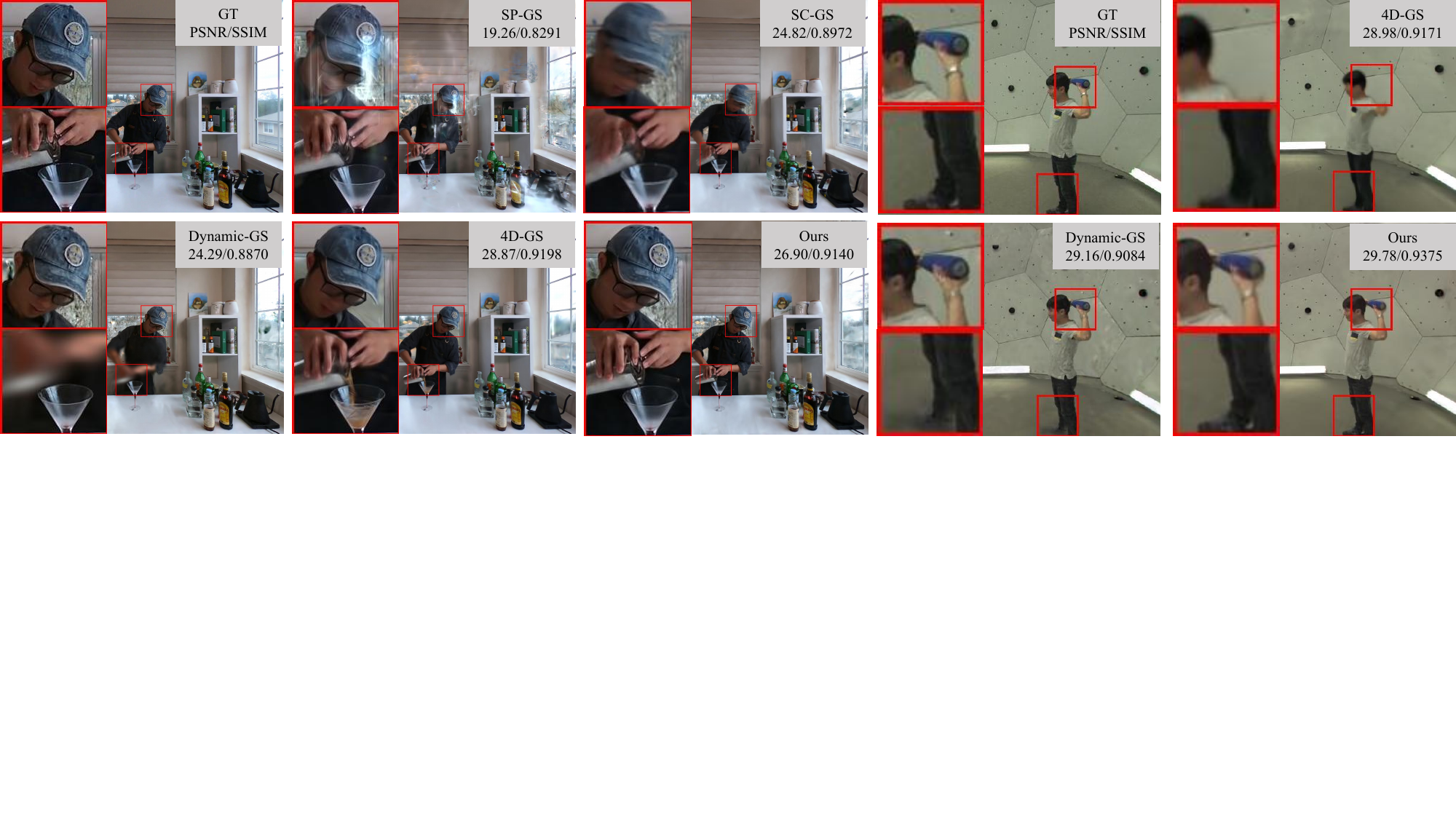}
  \caption{\textbf{Left:} Frame 20 of the "coffee\_martini" sequence from the Neu3DV dataset.  \textbf{Right:} Frame 74 of the "softball" sequence from the CMU-Panoptic dataset.}
  \label{fig:sub}
  \vspace{-10pt}
\end{figure*}

\subsection{Datasets and Implementation Details}

\begin{table*}[ht]
\centering
\caption{Per-scene results for the Neu3DV dataset. Each cell is color-coded to denote performance ranking: \colorbox{best}{best} for the top performance, \colorbox{second}{second} for the second best, and \colorbox{third}{third} for the third best.}
\small
\fontsize{7}{11.5}\selectfont
\setlength{\tabcolsep}{8pt} % 设置列间距为 5pt
\begin{tabular}{lccccccccccc}
\toprule
\multicolumn{3}{c}{Scene}      & \multicolumn{3}{c}{\textit{sear\_steak}}                                                        & \multicolumn{3}{c}{\textit{cook\_spinach}}                                                      & \multicolumn{3}{c}{\textit{cut\_roasted\_beef}}                                                 \\
\multicolumn{3}{c}{Metrics}    & \multicolumn{1}{c}{PSNR$\uparrow$}      & \multicolumn{1}{c}{SSIM$\uparrow$}       & \multicolumn{1}{c}{LPIPS$\downarrow$}      & \multicolumn{1}{c}{PSNR$\uparrow$}      & \multicolumn{1}{c}{SSIM$\uparrow$}       & \multicolumn{1}{c}{LPIPS$\downarrow$}      & \multicolumn{1}{c}{PSNR$\uparrow$}      & \multicolumn{1}{c}{SSIM$\uparrow$}       & \multicolumn{1}{c}{LPIPS$\downarrow$}      \\
\midrule
\multicolumn{3}{l}{Dynamic-GS~\numcite{luiten2023dynamic}} & 31.38                         & 0.9469                         & 0.1119                         & 29.98                         & 0.9388                         & 0.1179                         & 29.64                         & 0.9360                         & 0.1248                         \\
\multicolumn{3}{l}{MA-GS~\numcite{guo2024motion}} & 30.36                         & 0.9508                         & 0.0854                         & 31.15                         & 0.9378                         & 0.1053                         & 31.17                         & 0.9401                         & 0.1157                         \\
\multicolumn{3}{l}{4D-GS~\numcite{wu20234d}}      & 31.62                         & 0.9569                         & 0.0808                         & \cellcolor[HTML]{FFFC9E}32.79 & 0.9522                         & 0.0926                         & 32.13                         & 0.9467                         &   0.0959                       \\
\multicolumn{3}{l}{SP-GS~\numcite{wan2024superpoint}}      & 30.75                         & 0.9474                         & 0.0931                         & 31.32 & 0.9445                         & 0.0914                         & 30.44                         & 0.9457                         &   0.0942                       \\
\multicolumn{3}{l}{SC-GS~\numcite{huang2023sc}}      & 31.60                         & 0.9510                         & 0.1345                         & - & -                         & -                         & -                         & -                         &   -                       \\
\multicolumn{3}{l}{Ours-GoS1}  & \cellcolor[HTML]{FFFC9E}33.23 & \cellcolor[HTML]{FFFC9E}0.9654 & \cellcolor[HTML]{FFFC9E}0.0719 & \cellcolor[HTML]{F1A3A3}33.20 & \cellcolor[HTML]{F1A3A3}0.9586 & \cellcolor[HTML]{F1A3A3}0.0796 & \cellcolor[HTML]{F6CF88}33.00 & \cellcolor[HTML]{F1A3A3}0.9609 & \cellcolor[HTML]{F1A3A3}0.0795 \\
\multicolumn{3}{l}{Ours-GoS5}  & \cellcolor[HTML]{F1A3A3}33.72 & \cellcolor[HTML]{F1A3A3}0.9661 & \cellcolor[HTML]{F1A3A3}0.0704 & \cellcolor[HTML]{F6CF88}32.91 & \cellcolor[HTML]{F6CF88}0.9579 & \cellcolor[HTML]{F6CF88}0.0819 & \cellcolor[HTML]{F1A3A3}33.23 & \cellcolor[HTML]{F6CF88}0.9592 & \cellcolor[HTML]{F6CF88}0.0835 \\
\multicolumn{3}{l}{Ours-GoS10} & \cellcolor[HTML]{F6CF88}33.64 & \cellcolor[HTML]{F6CF88}0.9655 & \cellcolor[HTML]{F6CF88}0.0716 & 32.65                         & \cellcolor[HTML]{FFFC9E}0.9553 & \cellcolor[HTML]{FFFC9E}0.0861 & \cellcolor[HTML]{FFFC9E}32.47 & \cellcolor[HTML]{FFFC9E}0.9555 & \cellcolor[HTML]{FFFC9E}0.0890 \\
\midrule
\multicolumn{3}{c}{Scene}      & \multicolumn{3}{c}{\textit{flame\_steak}}                                                       & \multicolumn{3}{c}{\textit{flame\_salmon\_1}}                                                   & \multicolumn{3}{c}{\textit{coffee\_martini}}                                                    \\
\multicolumn{3}{c}{Metrics}    & \multicolumn{1}{c}{PSNR$\uparrow$}      & \multicolumn{1}{c}{SSIM$\uparrow$}       & \multicolumn{1}{c}{LPIPS$\downarrow$}      & \multicolumn{1}{c}{PSNR$\uparrow$}      & \multicolumn{1}{c}{SSIM$\uparrow$}       & \multicolumn{1}{c}{LPIPS$\downarrow$}      & \multicolumn{1}{c}{PSNR$\uparrow$}      & \multicolumn{1}{c}{SSIM$\uparrow$}       & \multicolumn{1}{c}{LPIPS$\downarrow$}      \\
\midrule
\multicolumn{3}{l}{Dynamic-GS~\numcite{luiten2023dynamic}} & 30.41                         & 0.9429                         & 0.1121                         & 20.19                         & 0.8875                         & 0.1583                         & 24.29                         & 0.8870                          & 0.1630                          \\
\multicolumn{3}{l}{MA-GS~\numcite{guo2024motion}} & 29.14                         & 0.9456                         & 0.1008                         & 25.05                         & 0.9075                         & 0.1274                         & 25.72                         & 0.8979                          & 0.1533                          \\

\multicolumn{3}{l}{4D-GS~\numcite{wu20234d}}      & 29.28                         & 0.9545                         & 0.0836                         & \cellcolor[HTML]{F1A3A3}28.27 & 0.9106                         & 0.1289                         & \cellcolor[HTML]{F1A3A3}28.87 & \cellcolor[HTML]{F1A3A3}0.9198 & \cellcolor[HTML]{F1A3A3}0.1168 \\
\multicolumn{3}{l}{SP-GS~\numcite{wan2024superpoint}}      & 25.59                         & 0.8934                         & 0.1248                         & 25.13 & 0.9057                         & 0.1320                         & 19.26                         & 0.8291                         &   0.1996                       \\
\multicolumn{3}{l}{SC-GS~\numcite{huang2023sc}}      & -                         & -                         & -                         & - & -                         & -                         & 24.82                         & 0.8972                         &   0.2239                       \\

\multicolumn{3}{l}{Ours-GoS1}  & \cellcolor[HTML]{FFFC9E}32.84 & \cellcolor[HTML]{F6CF88}0.9645 & \cellcolor[HTML]{F6CF88}0.0723 & \cellcolor[HTML]{F6CF88}28.00 & \cellcolor[HTML]{F1A3A3}0.9173 & \cellcolor[HTML]{F1A3A3}0.1083 & \cellcolor[HTML]{F6CF88}26.90  & \cellcolor[HTML]{F6CF88}0.9140  & \cellcolor[HTML]{F6CF88}0.1170  \\
\multicolumn{3}{l}{Ours-GoS5}  & \cellcolor[HTML]{F1A3A3}33.18 & \cellcolor[HTML]{F1A3A3}0.9649 & \cellcolor[HTML]{F1A3A3}0.0707 & \cellcolor[HTML]{FFFC9E}27.65 & \cellcolor[HTML]{F6CF88}0.9155 & \cellcolor[HTML]{F6CF88}0.1127 & \cellcolor[HTML]{FFFC9E}26.71 & \cellcolor[HTML]{FFFC9E}0.9119 & \cellcolor[HTML]{FFFC9E}0.1242 \\
\multicolumn{3}{l}{Ours-GoS10} & \cellcolor[HTML]{F6CF88}32.94 & \cellcolor[HTML]{FFFC9E}0.9631 & \cellcolor[HTML]{FFFC9E}0.0733  & 27.17                         & \cellcolor[HTML]{FFFC9E}0.9127 & \cellcolor[HTML]{FFFC9E}0.1165 & 26.51                         & 0.9100                           & 0.1283                      \\
\bottomrule
\end{tabular}
\label{Quantitative}
\end{table*}

\begin{table*}[ht]
\centering
\caption{Per-scene results for the CMU-Panoptic dataset.}
\fontsize{7}{11.5}\selectfont
\setlength{\tabcolsep}{8pt} 
\begin{tabular}{lccccccccccc}
\toprule
\multicolumn{3}{c}{Scene}      & \multicolumn{3}{c}{\textit{softball}}                                                        & \multicolumn{3}{c}{\textit{boxes}}                                                      & \multicolumn{3}{c}{\textit{basketball}}                                                 \\
\multicolumn{3}{c}{Metrics}    & \multicolumn{1}{c}{PSNR$\uparrow$}      & \multicolumn{1}{c}{SSIM$\uparrow$}       & \multicolumn{1}{c}{LPIPS$\downarrow$}      & \multicolumn{1}{c}{PSNR$\uparrow$}      & \multicolumn{1}{c}{SSIM$\uparrow$}       & \multicolumn{1}{c}{LPIPS$\downarrow$}      & \multicolumn{1}{c}{PSNR$\uparrow$}      & \multicolumn{1}{c}{SSIM$\uparrow$}       & \multicolumn{1}{c}{LPIPS$\downarrow$}      \\
\midrule
\multicolumn{3}{l}{Dynamic-GS~\numcite{luiten2023dynamic}} & 26.93                         & 0.9076                         & 0.1804                         & 27.79                         & 0.9069                         & 0.1769                         & \cellcolor[HTML]{F1A3A3}28.54                         & 0.9032                         & 0.1812                         \\
\multicolumn{3}{l}{Ours-GoS2}  & \cellcolor[HTML]{F1A3A3}27.48 & \cellcolor[HTML]{F1A3A3}0.9264 & \cellcolor[HTML]{F1A3A3}0.1374 & \cellcolor[HTML]{F1A3A3}27.88 & \cellcolor[HTML]{F1A3A3}0.9227 & \cellcolor[HTML]{F1A3A3}0.1413 & 27.72 & \cellcolor[HTML]{F1A3A3}0.9203 & \cellcolor[HTML]{F1A3A3}0.1423 \\
\bottomrule
\end{tabular}
\label{Quantitative2}
\vspace{-10pt}
\end{table*}

\paragraph{Neu3DV Dataset~\cite{li2021neural}.} The Neural 3D Video Synthesis Dataset includes six sequences, originally captured at a resolution of 2704 × 2028, which were downsampled to 1352 × 1014 for training. The sequence 'flame\_salmon\_1' contains 1200 frames, while the remaining five sequences consist of 300 frames each. All sequences were recorded using 15 to 20 static cameras, all placed to form a fanned-out arrangement in front of the scene.

% \vspace{-10pt} % 在 minipage 后添加

\paragraph{CMU-Panoptic Dataset~\cite{Joo_2017_TPAMI}.} The CMU-Panoptic Dataset comprises three sequences featuring complex, dynamic object motions. Each sequence is recorded at a resolution of 640 × 360 and consists of 150 frames. The data was collected using 31 static cameras evenly distributed in a spherical arrangement around the scene, with 27 used for training and 4 reserved for testing, 

\paragraph{Implementation Details.} 

All experiments were conducted on NVIDIA RTX 4070 GPUs. To evaluate the effectiveness of our H3D control point method for motion representation, we compared it with SP-GS~\cite{wan2024superpoint}, SC-GS~\cite{huang2023sc}, Dynamic-GS~\cite{luiten2023dynamic}, MA-GS~\cite{guo2024motion} and 4D-GS~\cite{wu20234d}, all maintaining a constant number of Gaussian points. Using the official implementations provided by each baseline, we ensured identical 3D point initialization. In accordance with the 4D-GS setup, we employed COLMAP~\cite{schoenberger2016sfm,schoenberger2016mvs} to initialize 3D points from the first training frames and followed the 10k-iteration reconstruction strategy of Dynamic-GS. For subsequent frames, we trained keyframes with 500 iterations and non-keyframes with 100 iterations, using a single Adam optimizer with fixed learning rates as in Dynamic-GS. 
Since our method requires both object masks and optical flow as inputs, we specify the sources for these data. For the Neu3DV Dataset, we obtained the object masks using the SAM-track method~\cite{cheng2023segment}. For the CMU Dataset, we reused the foreground masks provided by Dynamic-GS~\cite{luiten2023dynamic}. Unless otherwise specified, we used DIS~\cite{kroeger2016fast} as the default optical flow model.
The H3D control point sampling configurations were tailored to each dataset based on resolution and motion complexity. We set the grid interval to 64 pixels with a 16-pixel radius for the Neu3DV dataset, and 32 pixels with an 8-pixel radius for the CMU dataset.

\subsection{Reconstruction Results}
\paragraph{Quantitative Results.} 
We first report average quantitative results including PSNR, SSIM, and LPIPS in Tab.~\ref{Average}. Our method consistently outperformed SP-GS, MA-GS, Dynamic-GS, 4D-GS across all metrics and demonstrated a significant advantage in training time. While MA-GS also leveraged optical flow, it did so via a gradient-based method with motion represented as an implicit neural field. Our approach achieved superior visual quality and more efficient training.
Per-scene quantitative results are provided in Tab.~\ref{Quantitative} and Tab.~\ref{Quantitative2}. 
Our approach achieved state-of-the-art (SOTA) PSNR performance in most sequences and yields even better SSIM and LPIPS scores across more scenes, indicating improved perceptual reconstruction quality.
In the Neu3DV dataset, SC-GS demonstrates vulnerability when dealing with real-world dynamic scenarios, failing to converge efficiently across multiple sequences.
In the CMU-Panoptic dataset, MA-GS, SP-GS, SC-GS, 4D-GS all encounter difficulties handling dynamic motion. For instance, 4D-GS fails under rapid scene movement, resulting in object disappearance as depicted in Fig.~\ref{fig:sub}.

\begin{figure}[htbp]
\centering
\begin{minipage}[t]{0.45\textwidth}  % 左边放图
    \vspace{0pt}
    \centering
    \includegraphics[width=1\textwidth]{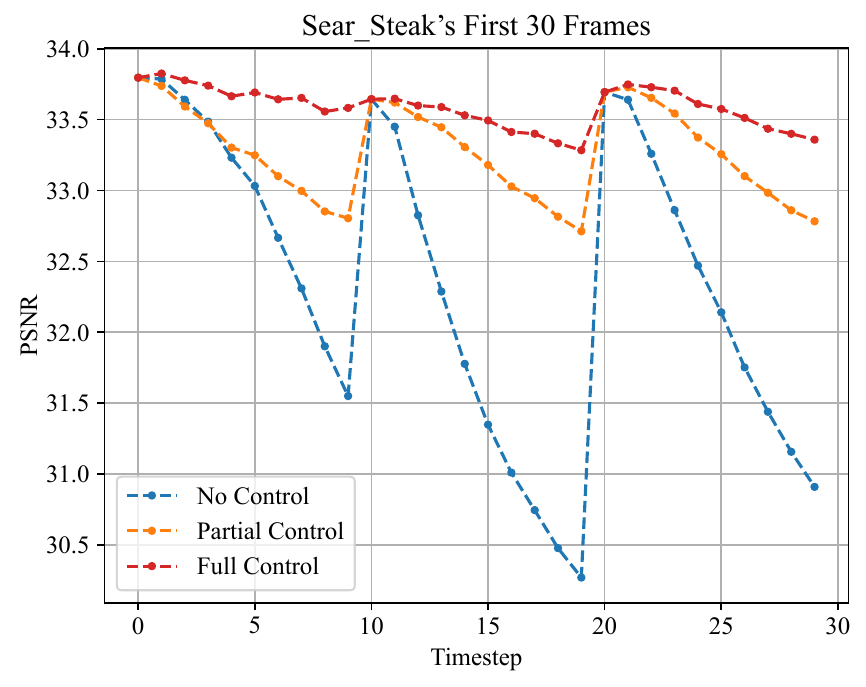}
    \captionof{figure}{An example illustrating reconstruction quality degradation across frames under three different settings of 3D control points.}
    \label{fig:MSE_plot}
\end{minipage}
\hfill
\begin{minipage}[t]{0.45\textwidth}  % 右边放两个表
    \centering
    \fontsize{7}{9.5}\selectfont
    \setlength{\tabcolsep}{6pt}
    \captionof{table}{Comparison between Gaussians and control points.}
    \label{tab:param}
    \begin{tabular}{lccc}
    \toprule
      \begin{tabular}[c]{@{}c@{}}Points\\ Category\end{tabular} 
    & \begin{tabular}[c]{@{}c@{}}Points\\ Num.\end{tabular} 
    & \begin{tabular}[c]{@{}c@{}}Attr.\\ Dim.\end{tabular} 
    & \begin{tabular}[c]{@{}c@{}}Param.\\ Num.\end{tabular}\\

    \midrule
    Scene GS          & $>100k$ & $\geq 13$ & $>1000k$ \\
    Obj. GS         & $\sim 10k$  & $\geq 13$ & $>100k$  \\
    Obj. Ctrl   & $0.2k-2.5k$ & $9$       & $1.8k-22.5k$ \\
    \bottomrule
    \end{tabular}
    
    \vspace{8pt}  % 控制两个表格之间的间距 
    \captionof{table}{Comparison between different optical flow methods.}
    \label{tab:cmp}
    \fontsize{7}{9.5}\selectfont
    \setlength{\tabcolsep}{5.5pt} % 设置列间距为 5pt
    \begin{tabularx}{\linewidth}{lcccc}
    \toprule
    O.F. Model & 2D MSE$\downarrow$ & PSNR$\uparrow$                          & SSIM$\uparrow$                             & LPIPS$\downarrow$                          \\
    \midrule
    PWC~\numcite{sun2018pwc}              &    \cellcolor[HTML]{FFFC9E}4.553e-5    & \cellcolor[HTML]{FFFC9E}33.39 & \cellcolor[HTML]{FFFC9E}0.9644 & \cellcolor[HTML]{FFFC9E}0.0737 \\
    SpyNet~\numcite{ranjan2017optical}             &   \cellcolor[HTML]{F6CF88}1.509e-5     & \cellcolor[HTML]{F6CF88}33.55 & \cellcolor[HTML]{F6CF88}0.9649 & \cellcolor[HTML]{F6CF88}0.0725 \\
    DIS~\numcite{kroeger2016fast}                &    \cellcolor[HTML]{EF949F}1.230e-5    & \cellcolor[HTML]{EF949F}33.64 & \cellcolor[HTML]{EF949F}0.9655 & \cellcolor[HTML]{EF949F}0.0716 \\
    \bottomrule
    \end{tabularx}
    
\end{minipage}
\vspace{-15pt}
\end{figure}

\paragraph{Subjective Assessment.}

In the Neu3DV dataset, the only case where our method did not surpass 4D-GS in objective metrics is shown in Fig.~\ref{fig:sub}. This is primarily due to suboptimal static scene initialization, which is decoupled from the subsequent motion modeling in our pipeline. The resulting background artifacts lowered the overall score. In dynamic regions, however, our method still delivered superior detail fidelity and accurate scene reconstruction.  In contrast, 4D-GS benefited from global optimization of the static background, boosting its overall metric but producing inferior quality in motion areas. Notably, 4D-GS reconstructed non-existent coffee liquid. SP-GS amplified background artifacts by using unreliable clustering centers as motion control points. SC-GS suffered from blurry dynamic regions,   as its high-degree-of-freedom control points are challenging to train via gradient descent.  Dynamic-GS produced blurry results because their complex loss functions lack generalization to dynamic scenarios.

In the CMU-Panoptic dataset, Dynamic-GS reconstructed noisy backgrounds and blurred feet, while 4D-GS struggled with large motion due to limitations in its global deformation field. 
Our method, by contrast, preserved fine textures and structural details without over-smoothing or object disappearance. Subjective comparisons clearly show our method excels in handling challenging motion. Further qualitative results are provided in the \textbf{supplementary video}.

\paragraph{Covergence Speed.} 
Our non-keyframes converged within 2 seconds per frame, and keyframes took approximately 10 seconds each. We anticipate further reductions in processing time as we continue refining the implementation.

\subsection{Ablation Study}

\paragraph{Points Parameter Comparison.} 

In Tab.~\ref{tab:param}, we compare 3D points across various categories to highlight the compactness of our control point method. Each Gaussian is characterized by 13 attributes. The number of parameters increases with higher harmonic degrees. In contrast, the number of 3D control points is significantly smaller, and their attributes remain fixed, indicating a much more compact representation and greater potential for streaming transmission.

\paragraph{Effectiveness of H3D Control Points.}

The rendering quality of three methods was evaluated under the GoS-10 configuration: \textcolor{blue}{\textbf{No Control}}, which lacks motion modeling and simply duplicates keyframes to subsequent non-keyframes; \textcolor{orange}{\textbf{Partial Control}}, which manipulates the scene using only the projected motion attributes in H3D control points; and \textcolor{red}{\textbf{Full Control}}, which manipulates the scene using the complete set of motion attributes in H3D control points.

The PSNR values for the first 30 frames of the sear\_steak sequence are presented in Fig.~\ref{fig:MSE_plot}. The \textcolor{blue}{\textbf{No Control}} method experienced rapid quality degradation over time, while \textcolor{orange}{\textbf{Partial Control}} showed moderate improvement by modeling partial motion. In contrast, \textcolor{red}{\textbf{Full Control}} consistently maintained higher quality throughout the sequence. 

By examining the gap between curves over each segment (e.g., steps 0–9, 10–19, 20–29), one can observe the progressive contribution of each component. Moreover, focusing solely on the \textcolor{red}{\textbf{Full Control}} setting, which corresponds to the GoS-10 setting in our main experiment, one can evaluate how reconstruction quality evolves over time in a streaming setup. As small errors accumulate frame-by-frame (e.g., 0–9, 10–19, 20–29), the residual compensation module applied at keyframes (e.g., 9–10, 19–20) effectively corrects these accumulated errors and restores high-quality reconstruction.

% Notably, Keyframe updates at timesteps 10 and 20 effectively mitigated accumulated errors. The observed PSNR gap underscores the benefit of integrating both projected and learned control point attributes, yielding substantial gains in reconstruction fidelity.

\paragraph{More Advanced Optical Flow Method Leads to Better Reconstruction Result.}

We further analyzed the impact of different optical flow algorithms~\cite{kroeger2016fast, ranjan2017optical, sun2018pwc} on reconstruction performance.  Among them, DIS~\cite{kroeger2016fast} provided the most accurate 2D flow estimates on the Neu3DV dataset. Optical flow accuracy was assessed via the average MSE between each frame and its warped predecessor, where lower 2D alignment error correlated strongly with improved 3D reconstruction quality. This highlights the potential for enhancing the pipeline by adopting more advanced optical flow methods. Furthermore, the observed positive correlation between optical flow accuracy and reconstruction quality validates the effectiveness of our 3D motion modeling approach.

%% file: sec/5_conclusion.tex
\section{Conclusion and Discussion}

We propose a novel heterogeneous 3D motion representation framework to address the challenges of dynamic scene reconstruction. By integrating discrete local motion modeling and leveraging H3D control points, our approach effectively decouples observable and learnable components of 3D motion, enabling precise and flexible representation. The framework further establishes a streaming pipeline that incorporates key innovations in 3D segmentation, motion manipulation, and residual compensation, ensuring robust and efficient reconstruction.

Our method surpasses existing state-of-the-art 4D Gaussian splatting approaches on real-world datasets. However, it still has limitations. The quality of 4D reconstruction is influenced by the initial static reconstruction, which remains an underexplored area with room for improvement. Additionally, the current method requires multi-view inputs from initial frames and does not support monocular video, a limitation we aim to address in future work.

Another limitation arises from background artifacts such as jittering, which are particularly noticeable in static regions (e.g., windows in some video sequences). These artifacts primarily stem from the incremental nature of our streaming reconstruction framework. Unlike global optimization methods that jointly optimize across all frames, our method updates each frame sequentially based on the previous reconstruction and the current input. While this design improves flexibility and enables low-latency streaming reconstruction, it also introduces susceptibility to cumulative noise and temporal drift, leading to observable floaters and jitter in the background. Addressing these artifacts presents an important direction for future work, for example by incorporating lightweight temporal smoothing or consistency mechanisms, while maintaining the efficiency of online reconstruction.

\begin{ack}
This work was supported by the National Natural Science Foundation of China under Grant Nos. 62431015 and 62471290.
\end{ack}

% Experimental results demonstrate the effectiveness of our method. The integration of both projected and learned attributes in H3D control points significantly enhances reconstruction quality, as evidenced by consistent improvements in rendering metrics across multiple configurations. Our approach achieves state-of-the-art performance on benchmark datasets such as Neu3DV and CMU-Panoptic, while maintaining low computational overhead and rapid convergence.
% In conclusion, the proposed method advances the field of dynamic 3D scene reconstruction by combining rigorous motion modeling with a structured streaming framework. Future research could explore extending this framework to handle more complex dynamic scenes and incorporating real-time capabilities for broader applicability.

% We introduce a novel discrete 6-DoF motion decoupling model that combines traditional graphics with learnable pipelines. This approach employs partially learnable control points for local 6-DoF motion representation, enabling fast convergence and robust reconstruction for real-world datasets. Additionally, we have developed an innovative workflow for streaming 4D real-world reconstruction using Gaussians and 3D control points. Starting with an initial 3D scene reconstruction, our approach progresses through several independent submodules, allowing each to be optimized individually for future improvements. 

%% file: sec/X_suppl.tex
\clearpage
\appendix

\section{Appendix}
\label{sec:rationale}
\subsection{Near-Parallel Light Hypothesis}
\label{Near-Parallel}

\begin{figure*}[ht]
  \centering
   \subfigure[]{
    \includegraphics[width=0.30\textwidth]{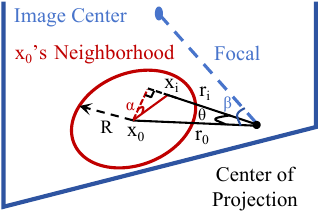}
    \label{fig:parallel2}
    }
    \subfigure[]{
    \includegraphics[width=0.65\linewidth]{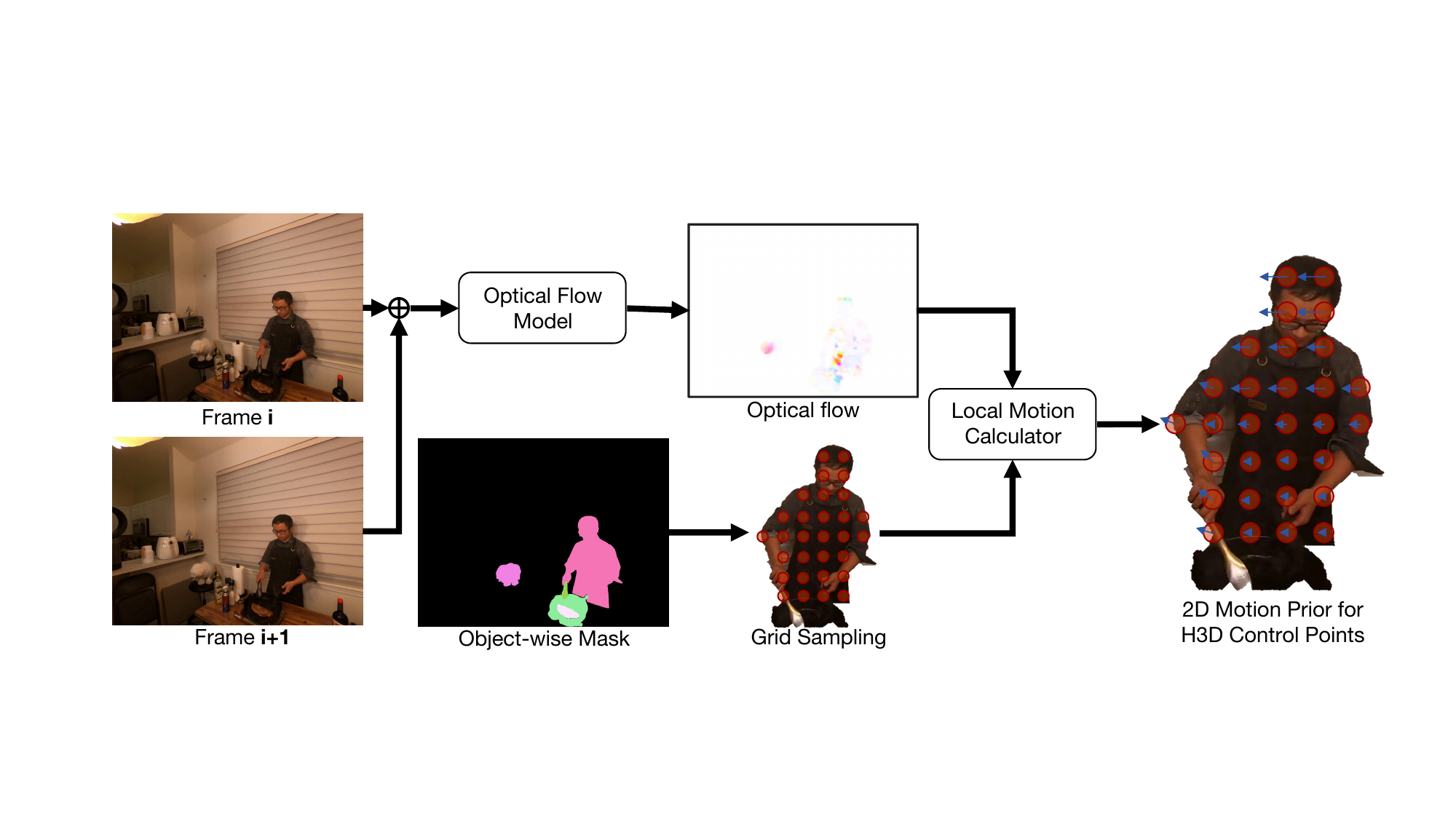}
    \label{fig:ops}
  }
  \caption{\textbf{Detailed diagrams for near-parallel light hypothesis proof and H3D control points sampling:} (a) Illustration for angles, points, rays in $\mathbf{x}_0$'s neighborhood. (b) Workflow for acquiring 2D motion prior. }
\end{figure*}

In the image plane, we consider a small region centered at $\mathbf{x}_0$ with radius $R$. Within this region, the rays from the camera projection center to each pixel can be approximated as nearly parallel. A detailed illustration is provided in Fig.~\ref{fig:parallel2}. For an arbitrary point $\mathbf{x}_i$ in the neighborhood of $\mathbf{x}_0$, we aim to show that the angle $\theta$ between the reference ray $\mathbf{r}_0$ (passing through $\mathbf{x}_0$) and the neighboring ray $\mathbf{r}_i$ is a first-order small quantity.

To begin, we drop a perpendicular (plumb line) from $\mathbf{x}_0$ to the ray $\mathbf{r}_i$, representing the shortest distance between the point and the ray. The length of this perpendicular segment can be expressed as:

\begin{equation}
\left\|\mathbf{x}_i-\mathbf{x}_0\right\| \cdot \cos \alpha,
\label{theta0}
\end{equation}

where $\left|\cdot\right|$ denotes the Euclidean distance, and $\alpha$ is the angle between the plumb line and the line connecting $\mathbf{x}_0$ to $\mathbf{x}_i$.

Next, applying the cosine theorem, we represent the distance from the projection center to $\mathbf{x}_0$ as

\begin{equation}
 f \cdot 1 / \cos \beta,
\label{theta1}
\end{equation}

where $f$ is the camera focal length, and $\beta$ is the angle between the ray $\mathbf{r}_0$ and the principal optical axis. Then, using the sine theorem, we can express the angle $\theta$ between rays $\mathbf{r}_0$ and $\mathbf{r}_i$:
\begin{equation}
    \theta_i=\mathbf{arcsin}(\frac{\left\|\mathbf{x}_i-\mathbf{x}_0\right\|}{f} \cdot \frac{\cos \alpha}{1 / \cos \beta}).
\label{theta}
\end{equation}

Since $\left\|\mathbf{x}_i-\mathbf{x}_0\right\|$ is always less than $R$, and the trigonometric terms involved are bounded by 1, the angle $\theta$ remains small as long as the focal length $f$ is significantly larger than the neighborhood radius $R$. 

Therefore, we conclude the proof of local ray near-parallelism. In practical applications, using a normalized focal length greater than 1000 pixels, it is sufficient to restrict the local region to a radius of 50 pixels to ensure the validity of this approximation.

\subsection{2D Motion Prior Acquisition}

We propose an intuitive workflow for acquiring 3D motion priors, as illustrated in Fig.~\ref{fig:ops}. The inputs are two consecutive frames and object-wise masks.

First, an optical flow network estimates dense motion between the two frames. we generate a sampling grid for each object within the view. This grid is then passed to the local motion calculator—an abstracted version of the method described in Sec.~\ref{Control}—which processes the grid and associates the resulting 2D motion priors with the corresponding 3D control points.

It is worth noting that the 3D control points are acquired independently from all training viewpoints. This multi-view acquisition results in a denser and more redundant distribution of control points compared to single-view methods, thereby reducing the effective influence range of each point. To compensate for this, we adopt a larger sampling interval when selecting control points and use a smaller neighborhood around each point for computing motion priors.

\begin{figure*}[ht]
  \centering
  \includegraphics[width=1\linewidth]{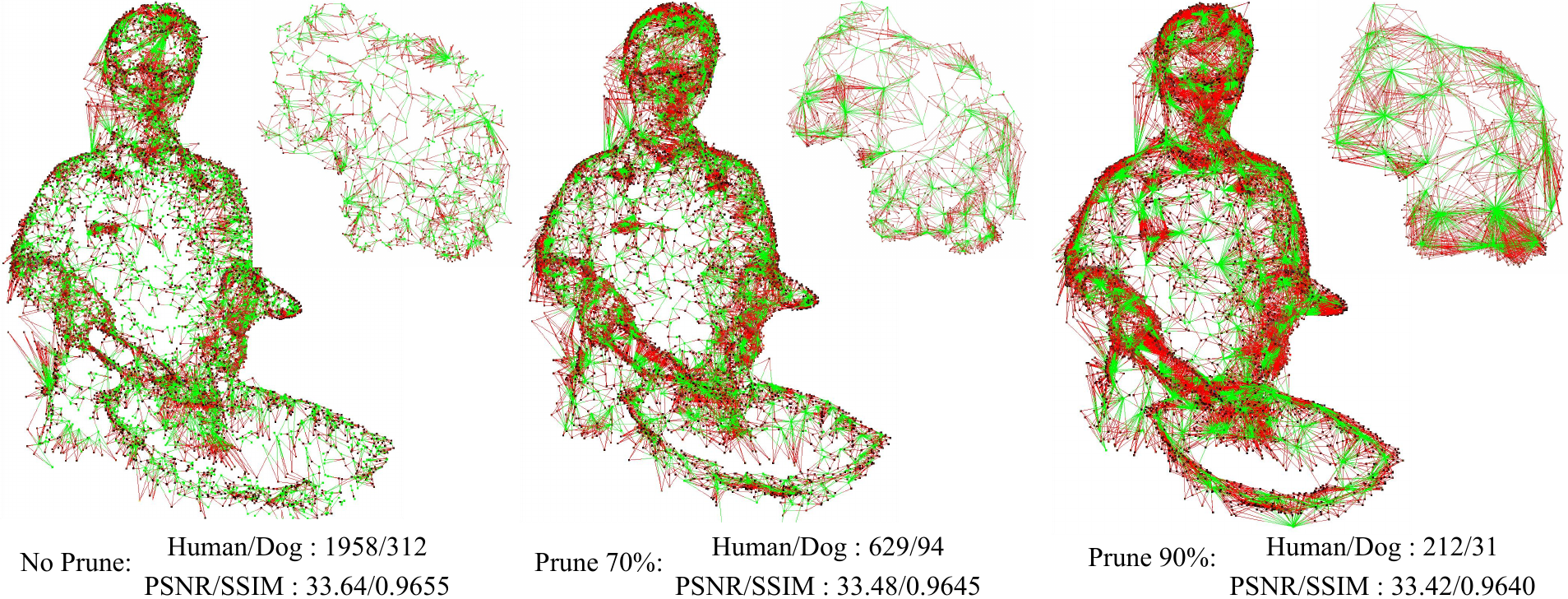}
  \caption{\textbf{Schematic comparison of Gaussians and control points for humans and dogs:} We visualized the topology using red and green line segments—red lines connect Gaussian points, while green lines connect control points. Additionally, we report the number of control points in the first frame for both human and dog subjects at various pruning rates, along with the corresponding reconstruction quality over the full sequence.}
  \label{fig:topology}
\end{figure*}

\subsection{H3D Control Points Sparsification}

H3D control point pruning is an optional step that further demonstrates the advantages of discrete motion representation. As shown in Fig.\ref{fig:topology}, we visualize the correspondence between Gaussians and control points under varying pruning rates. By sparsifying the control points, we achieve substantial gains in representation compactness while maintaining comparable reconstruction quality. For stable clustering initialization—especially when using a large number of cluster centers—we recommend adopting the k-means++ strategy\cite{arthur2007k}. This method offers a good trade-off between computational efficiency and reconstruction accuracy, making it well-suited for scenarios requiring densely distributed control points.

\subsection{3D Motion Visualization.}

\begin{figure}[ht]
  \centering
  \includegraphics[width=1\textwidth]{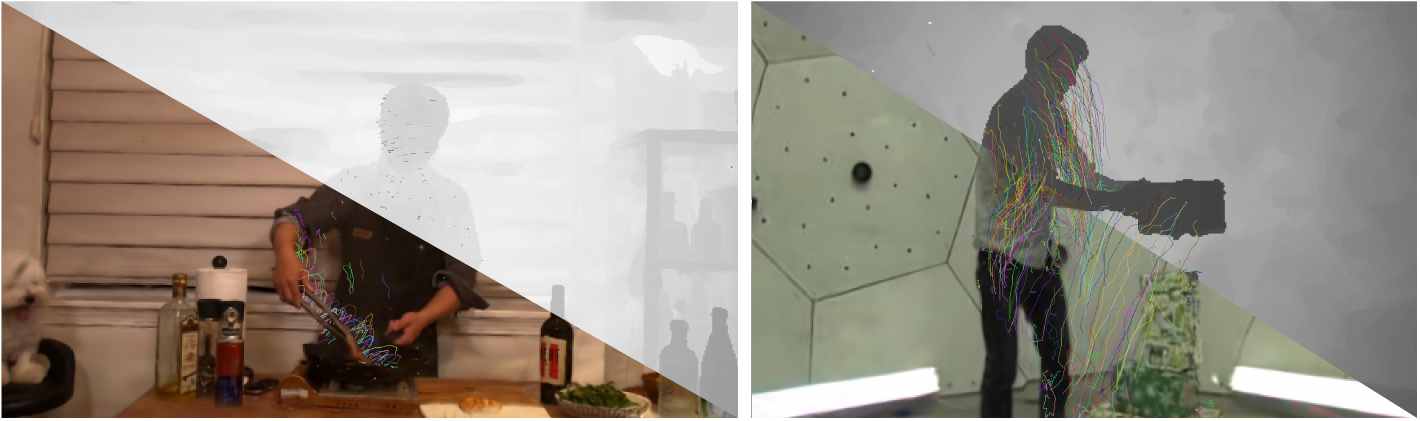}
  \caption{\textbf{3D Motion Visualization:} Visualization of Gaussian 3D motion  in the ``sear\_steak'' and ``boxes'' sequences.
  }
  \label{fig:sub3}
\end{figure}

We visualized the 3D motion of Gaussians in Fig.~\ref{fig:sub3}. The motion trajectories accurately depict the subject turning steaks and lifting boxes.

\begin{figure*}[ht]
  \centering
   \subfigure{
    \includegraphics[width=0.3\textwidth]{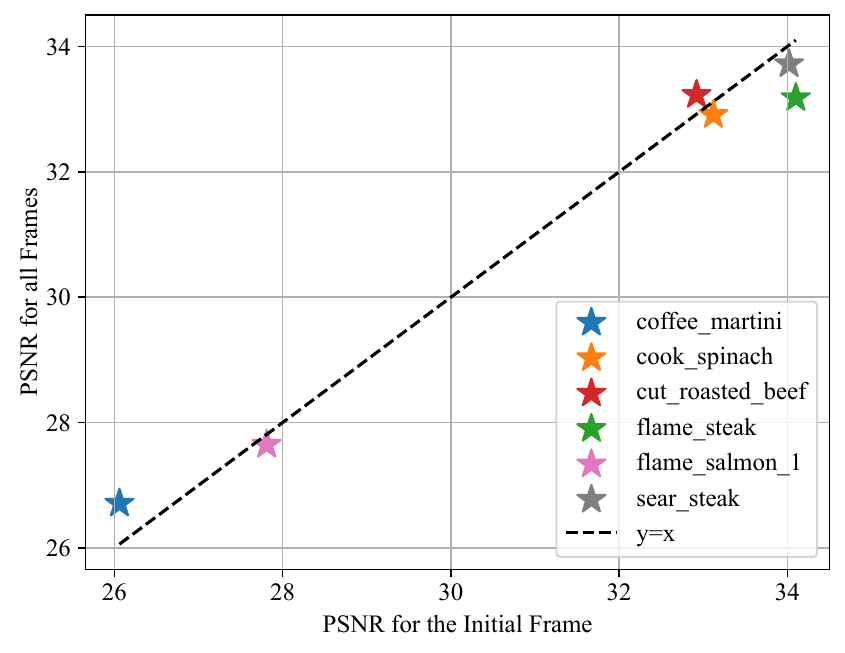}
    \label{fig:PSNR}
    }
    \subfigure{
    \includegraphics[width=0.3\textwidth]{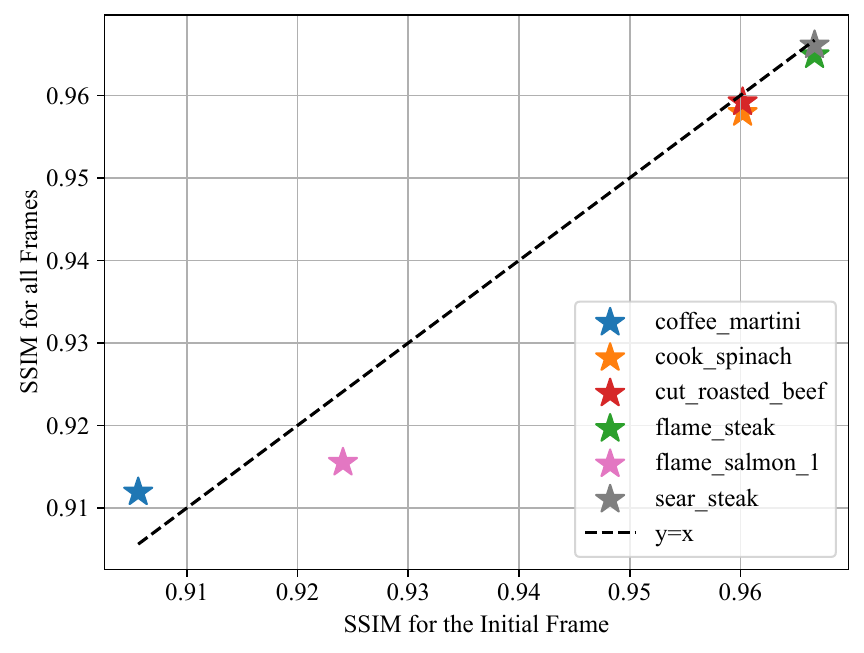}
    \label{fig:SSIM}
  }
  \subfigure{
    \includegraphics[width=0.3\textwidth]{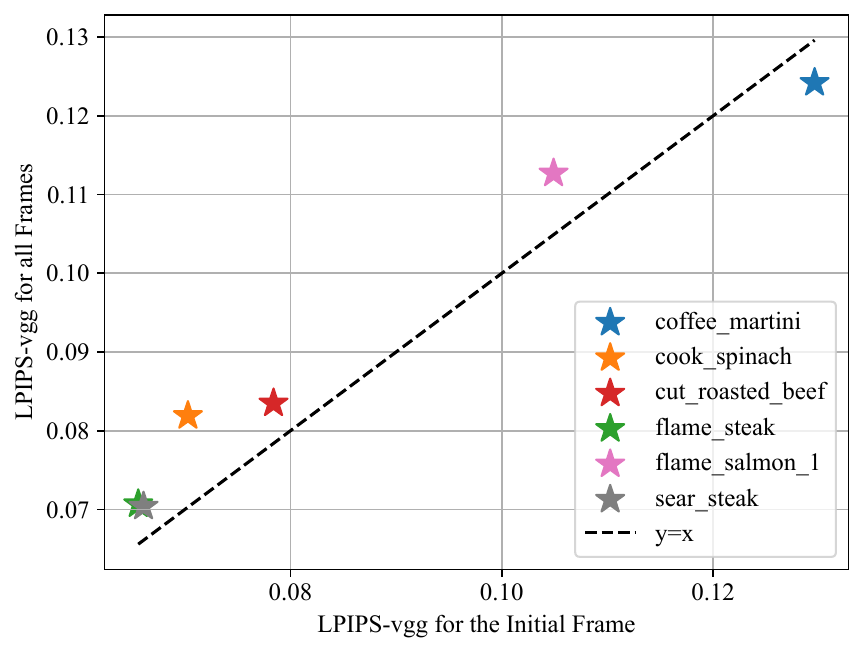}
    \label{fig:LPIPS}
  }
  \caption{Reconstruction quality correlation between the initial frame and the entire video.}
% \vspace{0.0cm}
  \label{fig:Metric}
\end{figure*}

\subsection{Significance of Initial 3D Scene.} 

The suboptimal static scene reconstruction in the first frame of the \textit{coffee\_martini} sequence noticeably impacted the quality of the subsequent 4D reconstruction. To further analyze this dependency, we evaluated PSNR, SSIM, and LPIPS-VGG metrics across sequences in the Neu3DV dataset. As shown in Fig.~\ref{fig:Metric}, the scatter plots exhibit a strong correlation between the quality of the initial 3D reconstruction and the final 4D results, with points closely following the y = x line. This alignment suggests that the initial static reconstruction defines an upper bound on the achievable dynamic reconstruction quality. Importantly, our method is designed to fully exploit this potential: it builds on the static scene without introducing new Gaussians during the sequence, making efficient use of the available representation regardless of its initial fidelity.

\subsection{More Detailed Settings for Fair Comparison}

\paragraph{SP-GS~\cite{wan2024superpoint} and SC-GS Setting.} We used the official codebases and initialized Gaussians from the same point cloud to ensure consistency.

\paragraph{MA-GS~\cite{guo2024motion} Setting.} We followed the official implementation and used the same initial point cloud. We selected the deformation-based pipeline, which employs an implicit neural network for motion representation and incorporates optical flow via gradient-based optimization.

\paragraph{Dynamic-GS~\cite{luiten2023dynamic} Setting.} This approach requires 2D foreground masks and segmentation labels for the initial 3D points. For a fair comparison, we merged our objects’ 2D masks and applied the labeling strategy described in Sec.~\ref{cate} to assign semantic categories to the initial 3D points. Additionally, we reduced the training iterations from 2k to 0.5k per frame when evaluating on the CMU-Panoptic dataset.

\paragraph{4D-GS~\cite{wu20234d} Setting.} For the ``flame\_salmon\_1'' sequence, four times longer than the other sequences, we expanded the training iterations from 17k to 68k to ensure a fair comparison.

\subsection{Additional Subjective Results at Novel Viewpoints}

We present additional qualitative results from various sequences in both image and video formats to support an intuitive evaluation of our method. The images are organized in two-row groups, arranged from left to right and top to bottom in the following order: GT, Dynamic-GS, 4D-GS, and Ours. Corresponding video results are provided in the accompanying MP4 files.

 % You can find them in our supplementary video. 

\begin{figure}[htbp]
    \centering
    \label{fig:subfigures}
    \caption{More subjective outputs.}
    \includegraphics[width=1\textwidth]{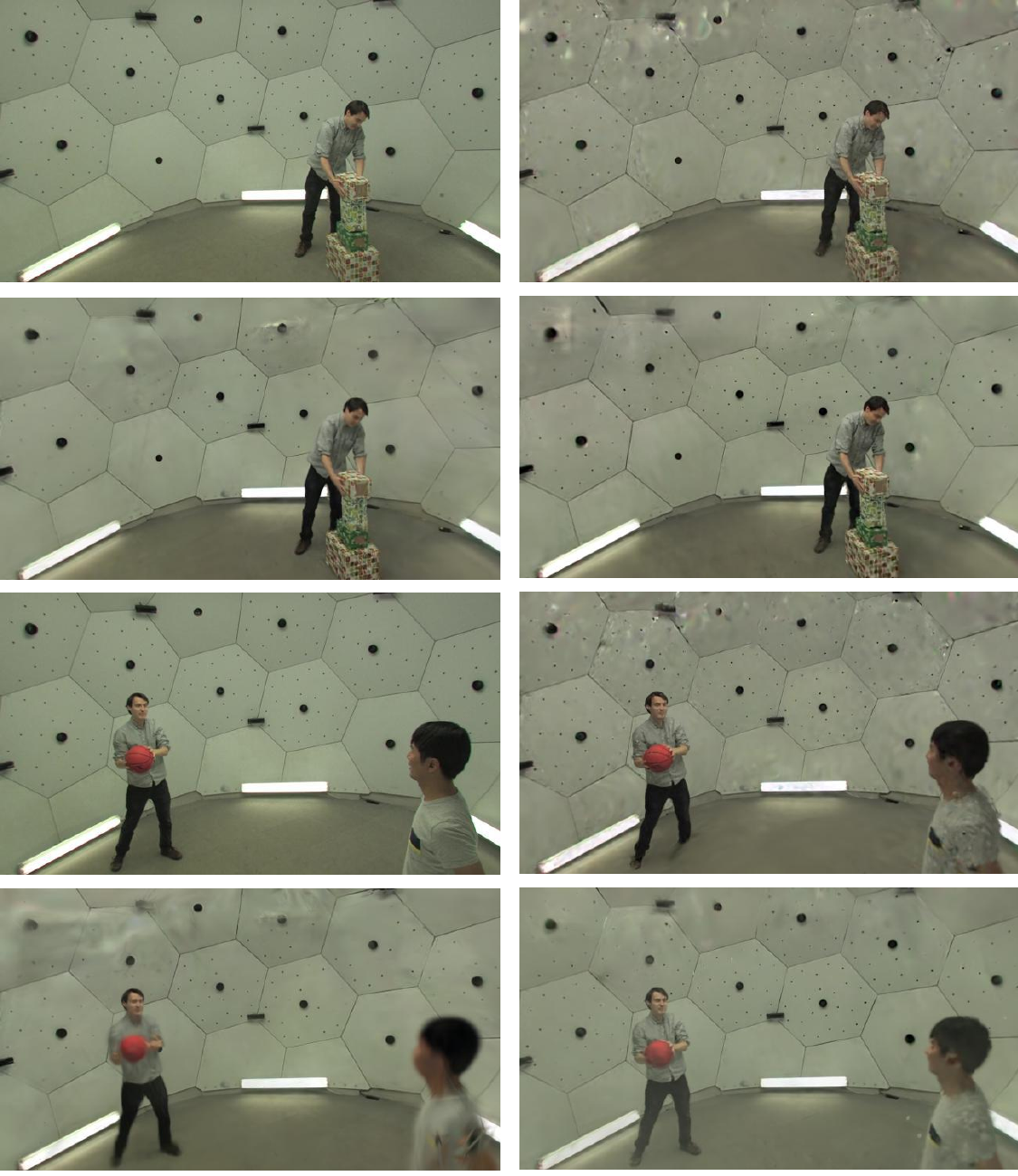}
\end{figure}

\begin{figure}[htbp]
    \centering
    \includegraphics[width=1\textwidth]{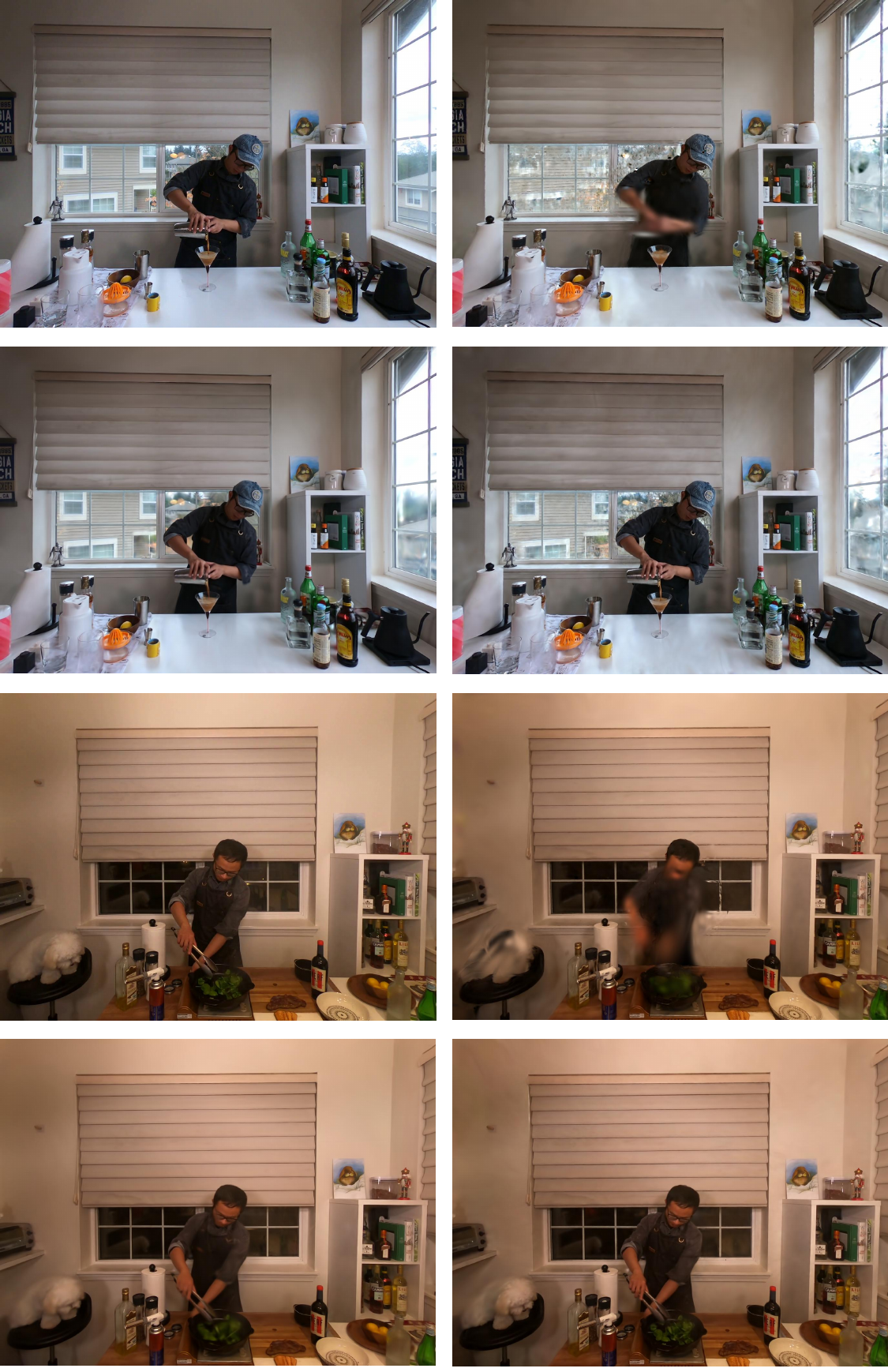}
\end{figure}

\begin{figure}[htbp]
    \centering

    \includegraphics[width=1\textwidth]{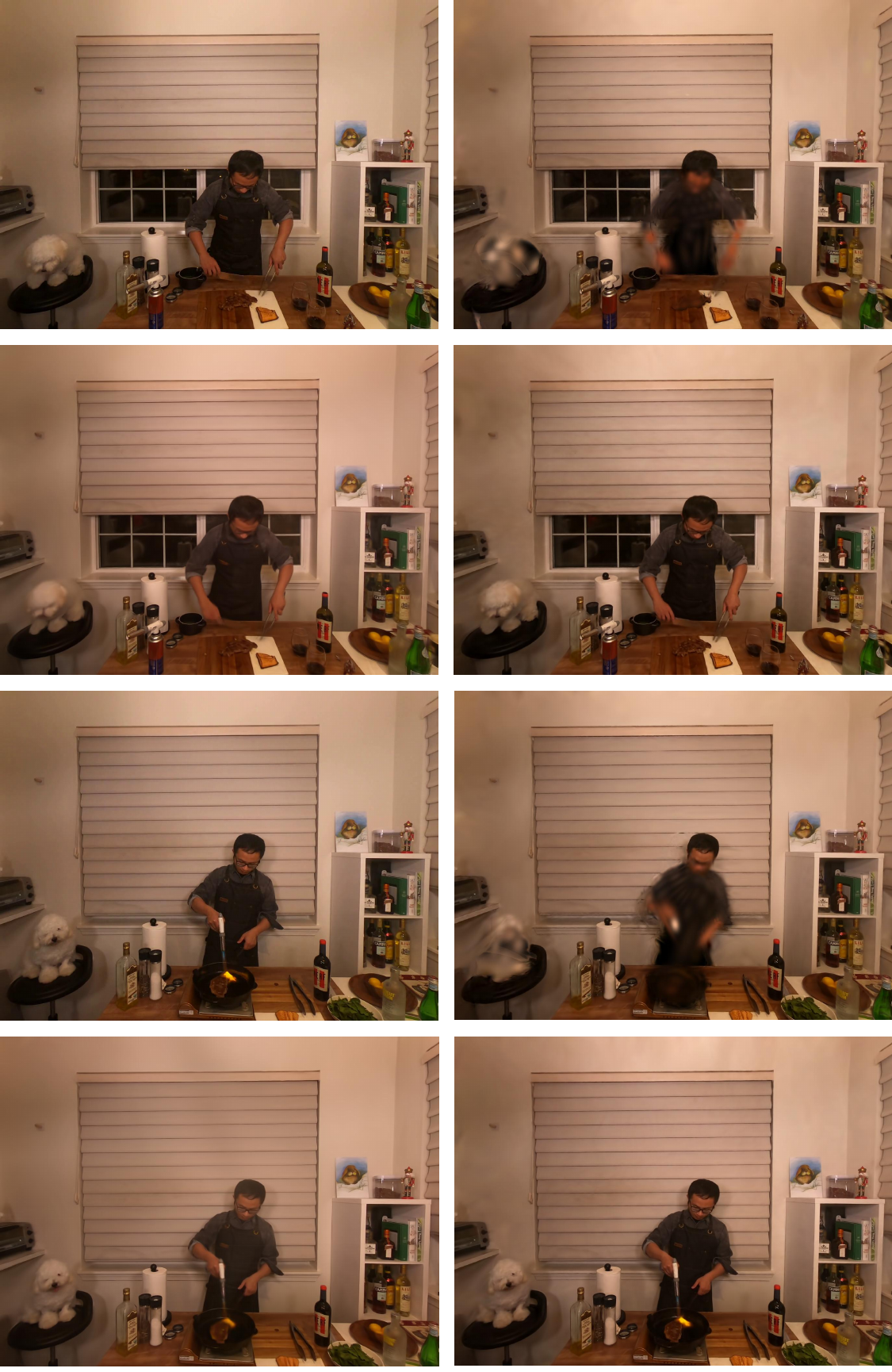}
\end{figure}

%% file: neurips_2025.bbl
\begin{thebibliography}{54}
\providecommand{\natexlab}[1]{#1}
\providecommand{\url}[1]{\texttt{#1}}
\expandafter\ifx\csname urlstyle\endcsname\relax
  \providecommand{\doi}[1]{doi: #1}\else
  \providecommand{\doi}{doi: \begingroup \urlstyle{rm}\Url}\fi

\bibitem[Arthur et~al.(2007)Arthur, Vassilvitskii, et~al.]{arthur2007k}
David Arthur, Sergei Vassilvitskii, et~al.
\newblock k-means++: The advantages of careful seeding.
\newblock In \emph{Soda}, pages 1027--1035, 2007.

\bibitem[Attal et~al.(2021)Attal, Laidlaw, Gokaslan, Kim, Richardt, Tompkin, and O'Toole]{attal2021torf}
Benjamin Attal, Eliot Laidlaw, Aaron Gokaslan, Changil Kim, Christian Richardt, James Tompkin, and Matthew O'Toole.
\newblock T{\"o}rf: Time-of-flight radiance fields for dynamic scene view synthesis.
\newblock \emph{Advances in neural information processing systems}, 34:\penalty0 26289--26301, 2021.

\bibitem[Cao and Johnson(2023)]{cao2023hexplane}
Ang Cao and Justin Johnson.
\newblock Hexplane: A fast representation for dynamic scenes.
\newblock In \emph{Proceedings of the IEEE/CVF Conference on Computer Vision and Pattern Recognition}, pages 130--141, 2023.

\bibitem[Cheng et~al.(2023)Cheng, Li, Xu, Li, Yang, Wang, and Yang]{cheng2023segment}
Yangming Cheng, Liulei Li, Yuanyou Xu, Xiaodi Li, Zongxin Yang, Wenguan Wang, and Yi Yang.
\newblock Segment and track anything.
\newblock \emph{arXiv preprint arXiv:2305.06558}, 2023.

\bibitem[Cover and Hart(1967)]{cover1967nearest}
Thomas Cover and Peter Hart.
\newblock Nearest neighbor pattern classification.
\newblock \emph{IEEE transactions on information theory}, 13\penalty0 (1):\penalty0 21--27, 1967.

\bibitem[Das et~al.(2024)Das, Wewer, Yunus, Ilg, and Lenssen]{das2024neural}
Devikalyan Das, Christopher Wewer, Raza Yunus, Eddy Ilg, and Jan~Eric Lenssen.
\newblock Neural parametric gaussians for monocular non-rigid object reconstruction.
\newblock In \emph{Proceedings of the IEEE/CVF Conference on Computer Vision and Pattern Recognition}, pages 10715--10725, 2024.

\bibitem[Du et~al.(2021)Du, Zhang, Yu, Tenenbaum, and Wu]{du2021neural}
Yilun Du, Yinan Zhang, Hong-Xing Yu, Joshua~B Tenenbaum, and Jiajun Wu.
\newblock Neural radiance flow for 4d view synthesis and video processing.
\newblock In \emph{2021 IEEE/CVF International Conference on Computer Vision (ICCV)}, pages 14304--14314. IEEE Computer Society, 2021.

\bibitem[Fang et~al.(2022)Fang, Yi, Wang, Xie, Zhang, Liu, Nie{\ss}ner, and Tian]{fang2022fast}
Jiemin Fang, Taoran Yi, Xinggang Wang, Lingxi Xie, Xiaopeng Zhang, Wenyu Liu, Matthias Nie{\ss}ner, and Qi Tian.
\newblock Fast dynamic radiance fields with time-aware neural voxels.
\newblock In \emph{SIGGRAPH Asia 2022 Conference Papers}, pages 1--9, 2022.

\bibitem[Fridovich-Keil et~al.(2023)Fridovich-Keil, Meanti, Warburg, Recht, and Kanazawa]{fridovich2023k}
Sara Fridovich-Keil, Giacomo Meanti, Frederik~Rahb{\ae}k Warburg, Benjamin Recht, and Angjoo Kanazawa.
\newblock K-planes: Explicit radiance fields in space, time, and appearance.
\newblock In \emph{Proceedings of the IEEE/CVF Conference on Computer Vision and Pattern Recognition}, pages 12479--12488, 2023.

\bibitem[Gao et~al.(2024)Gao, Xu, Cao, Mildenhall, Ma, Chen, Tang, and Neumann]{gao2024gaussianflow}
Quankai Gao, Qiangeng Xu, Zhe Cao, Ben Mildenhall, Wenchao Ma, Le Chen, Danhang Tang, and Ulrich Neumann.
\newblock Gaussianflow: Splatting gaussian dynamics for 4d content creation.
\newblock \emph{arXiv preprint arXiv:2403.12365}, 2024.

\bibitem[Guo et~al.(2024)Guo, Zhou, Li, Wang, and Li]{guo2024motion}
Zhiyang Guo, Wengang Zhou, Li Li, Min Wang, and Houqiang Li.
\newblock Motion-aware 3d gaussian splatting for efficient dynamic scene reconstruction.
\newblock \emph{IEEE Transactions on Circuits and Systems for Video Technology}, 2024.

\bibitem[Hu et~al.(2024)Hu, Wang, Fan, Fan, Peng, Lei, Li, and Zhang]{hu2024semantic}
Xu Hu, Yuxi Wang, Lue Fan, Junsong Fan, Junran Peng, Zhen Lei, Qing Li, and Zhaoxiang Zhang.
\newblock Semantic anything in 3d gaussians.
\newblock \emph{arXiv preprint arXiv:2401.17857}, 2024.

\bibitem[Huang et~al.(2024)Huang, Yu, Chen, Geiger, and Gao]{huang20242d}
Binbin Huang, Zehao Yu, Anpei Chen, Andreas Geiger, and Shenghua Gao.
\newblock 2d gaussian splatting for geometrically accurate radiance fields.
\newblock \emph{arXiv preprint arXiv:2403.17888}, 2024.

\bibitem[Huang et~al.(2023)Huang, Sun, Yang, Lyu, Cao, and Qi]{huang2023sc}
Yi-Hua Huang, Yang-Tian Sun, Ziyi Yang, Xiaoyang Lyu, Yan-Pei Cao, and Xiaojuan Qi.
\newblock Sc-gs: Sparse-controlled gaussian splatting for editable dynamic scenes.
\newblock \emph{arXiv preprint arXiv:2312.14937}, 2023.

\bibitem[Jiang et~al.(2023)Jiang, Shen, Wang, Su, Hong, Zhang, Yu, and Xu]{jiang2023hifi4g}
Yuheng Jiang, Zhehao Shen, Penghao Wang, Zhuo Su, Yu Hong, Yingliang Zhang, Jingyi Yu, and Lan Xu.
\newblock Hifi4g: High-fidelity human performance rendering via compact gaussian splatting.
\newblock \emph{arXiv preprint arXiv:2312.03461}, 2023.

\bibitem[Joo et~al.(2017)Joo, Simon, Li, Liu, Tan, Gui, Banerjee, Godisart, Nabbe, Matthews, Kanade, Nobuhara, and Sheikh]{Joo_2017_TPAMI}
Hanbyul Joo, Tomas Simon, Xulong Li, Hao Liu, Lei Tan, Lin Gui, Sean Banerjee, Timothy~Scott Godisart, Bart Nabbe, Iain Matthews, Takeo Kanade, Shohei Nobuhara, and Yaser Sheikh.
\newblock Panoptic studio: A massively multiview system for social interaction capture.
\newblock \emph{IEEE Transactions on Pattern Analysis and Machine Intelligence}, 2017.

\bibitem[Kerbl et~al.(2023)Kerbl, Kopanas, Leimk{\"u}hler, and Drettakis]{kerbl20233d}
Bernhard Kerbl, Georgios Kopanas, Thomas Leimk{\"u}hler, and George Drettakis.
\newblock 3d gaussian splatting for real-time radiance field rendering.
\newblock \emph{ACM Transactions on Graphics}, 42\penalty0 (4):\penalty0 1--14, 2023.

\bibitem[Kroeger et~al.(2016)Kroeger, Timofte, Dai, and Van~Gool]{kroeger2016fast}
Till Kroeger, Radu Timofte, Dengxin Dai, and Luc Van~Gool.
\newblock Fast optical flow using dense inverse search.
\newblock In \emph{Computer Vision--ECCV 2016: 14th European Conference, Amsterdam, The Netherlands, October 11--14, 2016, Proceedings, Part IV 14}, pages 471--488. Springer, 2016.

\bibitem[Li et~al.(2021)Li, Niklaus, Snavely, and Wang]{li2021neural}
Zhengqi Li, Simon Niklaus, Noah Snavely, and Oliver Wang.
\newblock Neural scene flow fields for space-time view synthesis of dynamic scenes.
\newblock In \emph{Proceedings of the IEEE/CVF Conference on Computer Vision and Pattern Recognition}, pages 6498--6508, 2021.

\bibitem[Li et~al.(2023)Li, Wang, Cole, Tucker, and Snavely]{li2023dynibar}
Zhengqi Li, Qianqian Wang, Forrester Cole, Richard Tucker, and Noah Snavely.
\newblock Dynibar: Neural dynamic image-based rendering.
\newblock In \emph{Proceedings of the IEEE/CVF Conference on Computer Vision and Pattern Recognition}, pages 4273--4284, 2023.

\bibitem[Li et~al.(2024)Li, Chen, Li, and Xu]{li2024spacetime}
Zhan Li, Zhang Chen, Zhong Li, and Yi Xu.
\newblock Spacetime gaussian feature splatting for real-time dynamic view synthesis.
\newblock In \emph{Proceedings of the IEEE/CVF Conference on Computer Vision and Pattern Recognition}, pages 8508--8520, 2024.

\bibitem[Lin et~al.(2022)Lin, Peng, Xu, Yan, Shuai, Bao, and Zhou]{lin2022efficient}
Haotong Lin, Sida Peng, Zhen Xu, Yunzhi Yan, Qing Shuai, Hujun Bao, and Xiaowei Zhou.
\newblock Efficient neural radiance fields for interactive free-viewpoint video.
\newblock In \emph{SIGGRAPH Asia 2022 Conference Papers}, pages 1--9, 2022.

\bibitem[Lin et~al.(2023)Lin, Peng, Xu, Xie, He, Bao, and Zhou]{lin2023im4d}
Haotong Lin, Sida Peng, Zhen Xu, Tao Xie, Xingyi He, Hujun Bao, and Xiaowei Zhou.
\newblock Im4d: High-fidelity and real-time novel view synthesis for dynamic scenes.
\newblock \emph{arXiv preprint arXiv:2310.08585}, 2023.

\bibitem[Lin et~al.(2024)Lin, Dai, Zhu, and Yao]{lin2024gaussian}
Youtian Lin, Zuozhuo Dai, Siyu Zhu, and Yao Yao.
\newblock Gaussian-flow: 4d reconstruction with dynamic 3d gaussian particle.
\newblock In \emph{Proceedings of the IEEE/CVF Conference on Computer Vision and Pattern Recognition}, pages 21136--21145, 2024.

\bibitem[Liu et~al.(2022)Liu, Cao, Mao, Zhang, Zhang, Keppo, Shan, Qie, and Shou]{liu2022devrf}
Jia-Wei Liu, Yan-Pei Cao, Weijia Mao, Wenqiao Zhang, David~Junhao Zhang, Jussi Keppo, Ying Shan, Xiaohu Qie, and Mike~Zheng Shou.
\newblock Devrf: Fast deformable voxel radiance fields for dynamic scenes.
\newblock \emph{Advances in Neural Information Processing Systems}, 35:\penalty0 36762--36775, 2022.

\bibitem[Liu et~al.(2023)Liu, Gao, Meuleman, Tseng, Saraf, Kim, Chuang, Kopf, and Huang]{liu2023robust}
Yu-Lun Liu, Chen Gao, Andreas Meuleman, Hung-Yu Tseng, Ayush Saraf, Changil Kim, Yung-Yu Chuang, Johannes Kopf, and Jia-Bin Huang.
\newblock Robust dynamic radiance fields.
\newblock In \emph{Proceedings of the IEEE/CVF Conference on Computer Vision and Pattern Recognition}, pages 13--23, 2023.

\bibitem[Luiten et~al.(2023)Luiten, Kopanas, Leibe, and Ramanan]{luiten2023dynamic}
Jonathon Luiten, Georgios Kopanas, Bastian Leibe, and Deva Ramanan.
\newblock Dynamic 3d gaussians: Tracking by persistent dynamic view synthesis.
\newblock \emph{arXiv preprint arXiv:2308.09713}, 2023.

\bibitem[Mildenhall et~al.(2021)Mildenhall, Srinivasan, Tancik, Barron, Ramamoorthi, and Ng]{mildenhall2021nerf}
Ben Mildenhall, Pratul~P Srinivasan, Matthew Tancik, Jonathan~T Barron, Ravi Ramamoorthi, and Ren Ng.
\newblock Nerf: Representing scenes as neural radiance fields for view synthesis.
\newblock \emph{Communications of the ACM}, 65\penalty0 (1):\penalty0 99--106, 2021.

\bibitem[Park et~al.(2021{\natexlab{a}})Park, Sinha, Barron, Bouaziz, Goldman, Seitz, and Martin-Brualla]{park2021nerfies}
Keunhong Park, Utkarsh Sinha, Jonathan~T Barron, Sofien Bouaziz, Dan~B Goldman, Steven~M Seitz, and Ricardo Martin-Brualla.
\newblock Nerfies: Deformable neural radiance fields.
\newblock In \emph{Proceedings of the IEEE/CVF International Conference on Computer Vision}, pages 5865--5874, 2021{\natexlab{a}}.

\bibitem[Park et~al.(2021{\natexlab{b}})Park, Sinha, Hedman, Barron, Bouaziz, Goldman, Martin-Brualla, and Seitz]{park2021hypernerf}
Keunhong Park, Utkarsh Sinha, Peter Hedman, Jonathan~T Barron, Sofien Bouaziz, Dan~B Goldman, Ricardo Martin-Brualla, and Steven~M Seitz.
\newblock Hypernerf: A higher-dimensional representation for topologically varying neural radiance fields.
\newblock \emph{arXiv preprint arXiv:2106.13228}, 2021{\natexlab{b}}.

\bibitem[Pumarola et~al.(2021)Pumarola, Corona, Pons-Moll, and Moreno-Noguer]{pumarola2021d}
Albert Pumarola, Enric Corona, Gerard Pons-Moll, and Francesc Moreno-Noguer.
\newblock D-nerf: Neural radiance fields for dynamic scenes.
\newblock In \emph{Proceedings of the IEEE/CVF Conference on Computer Vision and Pattern Recognition}, pages 10318--10327, 2021.

\bibitem[Ranjan and Black(2017)]{ranjan2017optical}
Anurag Ranjan and Michael~J Black.
\newblock Optical flow estimation using a spatial pyramid network.
\newblock In \emph{Proceedings of the IEEE conference on computer vision and pattern recognition}, pages 4161--4170, 2017.

\bibitem[Sch\"{o}nberger and Frahm(2016)]{schoenberger2016sfm}
Johannes~Lutz Sch\"{o}nberger and Jan-Michael Frahm.
\newblock Structure-from-motion revisited.
\newblock In \emph{Conference on Computer Vision and Pattern Recognition (CVPR)}, 2016.

\bibitem[Sch\"{o}nberger et~al.(2016)Sch\"{o}nberger, Zheng, Pollefeys, and Frahm]{schoenberger2016mvs}
Johannes~Lutz Sch\"{o}nberger, Enliang Zheng, Marc Pollefeys, and Jan-Michael Frahm.
\newblock Pixelwise view selection for unstructured multi-view stereo.
\newblock In \emph{European Conference on Computer Vision (ECCV)}, 2016.

\bibitem[Shao et~al.(2023)Shao, Zheng, Tu, Liu, Zhang, and Liu]{shao2023tensor4d}
Ruizhi Shao, Zerong Zheng, Hanzhang Tu, Boning Liu, Hongwen Zhang, and Yebin Liu.
\newblock Tensor4d: Efficient neural 4d decomposition for high-fidelity dynamic reconstruction and rendering.
\newblock In \emph{Proceedings of the IEEE/CVF Conference on Computer Vision and Pattern Recognition}, pages 16632--16642, 2023.

\bibitem[Song et~al.(2023)Song, Chen, Li, Chen, Chen, Yuan, Xu, and Geiger]{song2023nerfplayer}
Liangchen Song, Anpei Chen, Zhong Li, Zhang Chen, Lele Chen, Junsong Yuan, Yi Xu, and Andreas Geiger.
\newblock Nerfplayer: A streamable dynamic scene representation with decomposed neural radiance fields.
\newblock \emph{IEEE Transactions on Visualization and Computer Graphics}, 29\penalty0 (5):\penalty0 2732--2742, 2023.

\bibitem[Sorkine(2005)]{sorkine2005laplacian}
Olga Sorkine.
\newblock Laplacian mesh processing.
\newblock \emph{Eurographics (State of the Art Reports)}, 4\penalty0 (4):\penalty0 1, 2005.

\bibitem[Sumner et~al.(2007)Sumner, Schmid, and Pauly]{sumner2007embedded}
Robert~W Sumner, Johannes Schmid, and Mark Pauly.
\newblock Embedded deformation for shape manipulation.
\newblock In \emph{ACM siggraph 2007 papers}, pages 80--es. 2007.

\bibitem[Sun et~al.(2018)Sun, Yang, Liu, and Kautz]{sun2018pwc}
Deqing Sun, Xiaodong Yang, Ming-Yu Liu, and Jan Kautz.
\newblock Pwc-net: Cnns for optical flow using pyramid, warping, and cost volume.
\newblock In \emph{Proceedings of the IEEE conference on computer vision and pattern recognition}, pages 8934--8943, 2018.

\bibitem[Sun et~al.(2024)Sun, Jiao, Li, Zhang, Zhao, and Xing]{sun20243dgstream}
Jiakai Sun, Han Jiao, Guangyuan Li, Zhanjie Zhang, Lei Zhao, and Wei Xing.
\newblock 3dgstream: On-the-fly training of 3d gaussians for efficient streaming of photo-realistic free-viewpoint videos.
\newblock In \emph{Proceedings of the IEEE/CVF Conference on Computer Vision and Pattern Recognition}, pages 20675--20685, 2024.

\bibitem[Sze et~al.(2014)Sze, Budagavi, and Sullivan]{sze2014high}
Vivienne Sze, Madhukar Budagavi, and Gary~J Sullivan.
\newblock High efficiency video coding (hevc).
\newblock In \emph{Integrated circuit and systems, algorithms and architectures}, page~40. Springer, 2014.

\bibitem[Vedula et~al.(1999)Vedula, Baker, Rander, Collins, and Kanade]{vedula1999three}
Sundar Vedula, Simon Baker, Peter Rander, Robert Collins, and Takeo Kanade.
\newblock Three-dimensional scene flow.
\newblock In \emph{Proceedings of the Seventh IEEE International Conference on Computer Vision}, pages 722--729. IEEE, 1999.

\bibitem[Wan et~al.(2024)Wan, Lu, and Zeng]{wan2024superpoint}
Diwen Wan, Ruijie Lu, and Gang Zeng.
\newblock Superpoint gaussian splatting for real-time high-fidelity dynamic scene reconstruction.
\newblock \emph{arXiv preprint arXiv:2406.03697}, 2024.

\bibitem[Wang et~al.(2021)Wang, Eckart, Lucey, and Gallo]{wang2021neural}
Chaoyang Wang, Ben Eckart, Simon Lucey, and Orazio Gallo.
\newblock Neural trajectory fields for dynamic novel view synthesis.
\newblock \emph{arXiv preprint arXiv:2105.05994}, 2021.

\bibitem[Wang et~al.(2023{\natexlab{a}})Wang, MacDonald, Jeni, and Lucey]{wang2023flow}
Chaoyang Wang, Lachlan~Ewen MacDonald, Laszlo~A Jeni, and Simon Lucey.
\newblock Flow supervision for deformable nerf.
\newblock In \emph{Proceedings of the IEEE/CVF Conference on Computer Vision and Pattern Recognition}, pages 21128--21137, 2023{\natexlab{a}}.

\bibitem[Wang et~al.(2023{\natexlab{b}})Wang, Tan, Li, Tian, Song, and Liu]{wang2023mixed}
Feng Wang, Sinan Tan, Xinghang Li, Zeyue Tian, Yafei Song, and Huaping Liu.
\newblock Mixed neural voxels for fast multi-view video synthesis.
\newblock In \emph{Proceedings of the IEEE/CVF International Conference on Computer Vision}, pages 19706--19716, 2023{\natexlab{b}}.

\bibitem[Wiegand et~al.(2003)Wiegand, Sullivan, Bjontegaard, and Luthra]{wiegand2003overview}
Thomas Wiegand, Gary~J Sullivan, Gisle Bjontegaard, and Ajay Luthra.
\newblock Overview of the h. 264/avc video coding standard.
\newblock \emph{IEEE Transactions on circuits and systems for video technology}, 13\penalty0 (7):\penalty0 560--576, 2003.

\bibitem[Wu et~al.(2023)Wu, Yi, Fang, Xie, Zhang, Wei, Liu, Tian, and Wang]{wu20234d}
Guanjun Wu, Taoran Yi, Jiemin Fang, Lingxi Xie, Xiaopeng Zhang, Wei Wei, Wenyu Liu, Qi Tian, and Xinggang Wang.
\newblock 4d gaussian splatting for real-time dynamic scene rendering.
\newblock \emph{arXiv preprint arXiv:2310.08528}, 2023.

\bibitem[Xian et~al.(2021)Xian, Huang, Kopf, and Kim]{xian2021space}
Wenqi Xian, Jia-Bin Huang, Johannes Kopf, and Changil Kim.
\newblock Space-time neural irradiance fields for free-viewpoint video.
\newblock In \emph{Proceedings of the IEEE/CVF Conference on Computer Vision and Pattern Recognition}, pages 9421--9431, 2021.

\bibitem[Yang et~al.(2023{\natexlab{a}})Yang, Gao, Zhou, Jiao, Zhang, and Jin]{yang2023deformable}
Ziyi Yang, Xinyu Gao, Wen Zhou, Shaohui Jiao, Yuqing Zhang, and Xiaogang Jin.
\newblock Deformable 3d gaussians for high-fidelity monocular dynamic scene reconstruction.
\newblock \emph{arXiv preprint arXiv:2309.13101}, 2023{\natexlab{a}}.

\bibitem[Yang et~al.(2023{\natexlab{b}})Yang, Yang, Pan, Zhu, and Zhang]{yang2023real}
Zeyu Yang, Hongye Yang, Zijie Pan, Xiatian Zhu, and Li Zhang.
\newblock Real-time photorealistic dynamic scene representation and rendering with 4d gaussian splatting.
\newblock \emph{arXiv preprint arXiv:2310.10642}, 2023{\natexlab{b}}.

\bibitem[Yu et~al.(2024)Yu, Julin, Milacski, Niinuma, and Jeni]{yu2024cogs}
Heng Yu, Joel Julin, Zolt{\'a}n~{\'A} Milacski, Koichiro Niinuma, and L{\'a}szl{\'o}~A Jeni.
\newblock Cogs: Controllable gaussian splatting.
\newblock In \emph{Proceedings of the IEEE/CVF Conference on Computer Vision and Pattern Recognition}, pages 21624--21633, 2024.

\bibitem[Yu et~al.(2004)Yu, Zhou, Xu, Shi, Bao, Guo, and Shum]{yu2004mesh}
Yizhou Yu, Kun Zhou, Dong Xu, Xiaohan Shi, Hujun Bao, Baining Guo, and Heung-Yeung Shum.
\newblock Mesh editing with poisson-based gradient field manipulation.
\newblock In \emph{ACM SIGGRAPH 2004 Papers}, pages 644--651. 2004.

\bibitem[Zhu et~al.(2024)Zhu, Liang, Chang, Deng, Lu, Yang, Zhang, and Zhang]{zhu2024motiongs}
Ruijie Zhu, Yanzhe Liang, Hanzhi Chang, Jiacheng Deng, Jiahao Lu, Wenfei Yang, Tianzhu Zhang, and Yongdong Zhang.
\newblock Motiongs: Exploring explicit motion guidance for deformable 3d gaussian splatting.
\newblock \emph{Advances in Neural Information Processing Systems}, 37:\penalty0 101790--101817, 2024.

\end{thebibliography}
